%% file: main.tex
\documentclass[11pt, a4paper]{lumia}

\usepackage[sort&compress]{natbib}
\input{notations}

\usepackage{graphicx}           
\usepackage{tikz}               
\usepackage[edges]{forest}      

\usepackage{url}                
\usepackage{xurl}               

\usepackage{array}              
\usepackage{longtable}          
\usepackage{multirow}           
\usepackage{makecell}           
\usepackage{ragged2e}           

\usepackage{mathtools}          
\usepackage{nicefrac}           

\usepackage{algorithm}          
\usepackage{algorithmicx}       
\usepackage{algpseudocode}      
\usepackage{listings}           

\usepackage{subcaption}         
\usepackage{wrapfig}            
\usepackage[export]{adjustbox}  

\usepackage[commandnameprefix=ifneeded]{changes} 
\usepackage{xspace}             
\usepackage[normalem]{ulem}     
\usepackage{CJKutf8}            

\usepackage[tikz]{bclogo}       
\usepackage[framemethod=tikz]{mdframed} 

\usepackage{lipsum}             
\usepackage{tocloft}            
\usepackage{afterpage}          
\usepackage{bbding}             
\usepackage{epigraph}           
\usepackage{minitoc}            
\usepackage{multicol}           
\usepackage{textgreek}          

\newcolumntype{P}[1]{>{\RaggedRight\arraybackslash}p{#1}}

\definecolor{uclablue}{RGB}{39, 116, 174}
\definecolor{bigaired}{RGB}{156, 0, 0}
\definecolor{myblue}{HTML}{598BE7}
\definecolor{mildblue}{RGB}{31,119,180}
\definecolor{sectionblue}{RGB}{70, 130, 180}
\definecolor{methodblue}{RGB}{0, 150, 136}
\definecolor{bgblue}{RGB}{245,243,253}
\definecolor{ttblue}{RGB}{91,194,224}
\definecolor{mygreen}{rgb}{0.64, 0.56, 0.88}
\definecolor{myyellow}{rgb}{0.68, 0.6, 0.1}
\definecolor{fancygreen}{rgb}{0.33, 0.68, 0.20}
\definecolor{salmon}{rgb}{0.94, 0.52, 0.49}
\definecolor{tablegreen}{rgb}{0.82, 0.94, 0.75}
\definecolor{tableblue}{rgb}{0.81, 0.90, 0.94}
\definecolor{tablered}{rgb}{0.97, 0.85, 0.85}
\definecolor{tableorange}{rgb}{0.96, 0.85, 0.81}
\definecolor{myorange}{rgb}{1.0, 0.49, 0.0}
\definecolor{tlgreen}{rgb}{0.33, 0.68, 0.20}
\definecolor{darkgreen}{RGB}{0,100,0}
\definecolor{darkred}{RGB}{200, 0, 0}
\definecolor{customyellow}{HTML}{FFFACD}
\definecolor{refinegreen}{RGB}{0, 128, 75}
\definecolor{scoregreen}{RGB}{34, 139, 34}
\definecolor{hidden-blue}{RGB}{194,232,247}
\definecolor{hidden-black}{RGB}{20,68,106}
\definecolor{yes}{HTML}{C6EFCE}
\definecolor{no}{HTML}{FFC7CE}
\definecolor{partial}{HTML}{FFEB9C}
\definecolor{external}{HTML}{D9E1F2}
\definecolor{hdr}{HTML}{F2F2F2}
\definecolor{GRPOrow}{gray}{0.96}
\definecolor{FlowRLrow}{RGB}{225,236,255}
\definecolor{FlowBlue}{RGB}{80,120,210}
\definecolor{GRPOGray}{gray}{0.35}

\hypersetup{
    colorlinks=true, 
    citecolor=uclablue, 
    linkcolor=bigaired,
    urlcolor=darkblue
}

\setlist[itemize]{leftmargin=20pt, noitemsep, topsep=0pt}

\NewDocumentCommand{\kaiyan}{mO{}}{\textcolor{purple}{\textsuperscript{\textit{kaiyan}}\textsf{\textbf{\small[#1]}}}}
\NewDocumentCommand{\yuxin}{mO{}}{\textcolor{cyan}{\textsuperscript{\textit{yuxin}}\textsf{\textbf{\small[#1]}}}}
\NewDocumentCommand{\bx}{mO{}}{\textcolor{green}{\textsuperscript{\textit{bx}}\textsf{\textbf{\small[#1]}}}}
\NewDocumentCommand{\at}{mO{}}{\textcolor{red}{\textsuperscript{\textit{AT}}\textsf{\textbf{\small[#1]}}}}
\NewDocumentCommand{\re}{mO{}}{\textcolor{blue}{\textsuperscript{\textit{RE}}\textsf{\textbf{\small[#1]}}}}
\NewDocumentCommand{\ybsun}{mO{}}{\textcolor{magenta}{\textsuperscript{\textit{youbang}}\textsf{\textbf{\small[#1]}}}}
\NewDocumentCommand{\runze}{mO{}}{\textcolor{orange}{\textsuperscript{\textit{runze}}\textsf{\textbf{\small[#1]}}}}
\NewDocumentCommand{\add}{mO{}}{\textcolor{darkgreen}{\textsuperscript{\textit{Maybe Consider Discuss}}\textsf{\textbf{[#1]}}}}

\newcommand{\cmark}{\textcolor{darkgreen}{\boldmath$\checkmark$}}
\newcommand{\xmark}{\textcolor{darkred}{\boldmath$\times$}}

\newenvironment{itemize*}%
 {\leftmargini=10pt\begin{itemize}%
  \setlength{\itemsep}{0pt}%
  \setlength{\parskip}{0pt}%
  }%
 {\end{itemize}}

\newenvironment{enumerate*}%
 {\begin{enumerate}%
  \setlength{\itemsep}{0pt}%
  \setlength{\parskip}{0pt}}%
 {\end{enumerate}}

\newcommand{\cellstatus}[1]{%
  \begingroup
  \StrTrim{#1}[\statusval]%
  \IfStrEq{\statusval}{Yes}{\cellcolor{yes}\cmark}{}%
  \IfStrEq{\statusval}{No}{\cellcolor{no}\xmark}{}%
  \IfBeginWith{\statusval}{Yes (}{\cellcolor{yes}\cmark~\textit{\statusval\unskip}}{}%
  \IfStrEq{\statusval}{Partial}{\cellcolor{partial}\textbf{Partial}}{}%
  \IfStrEq{\statusval}{External}{\cellcolor{external}\textbf{External}}{}%
  \endgroup
}

\newtcolorbox{myboxi}[1][]{
  breakable,
  title=#1,
  colback=red!5,
  colbacktitle=red!5,
  coltitle=black,
  fonttitle=\bfseries,
  bottomrule=0pt,
  toprule=0pt,
  leftrule=2pt,
  rightrule=2pt,
  titlerule=0pt,
  arc=0pt,
  outer arc=0pt,
  colframe=red,
}

\newtcolorbox{myboxnote}[1][]{
  breakable,
  title=#1,
  colback=orange!0,
  colbacktitle=orange!0,
  coltitle=black,
  fonttitle=\bfseries,
  bottomrule=0pt,
  toprule=0pt,
  leftrule=2pt,
  rightrule=2pt,
  titlerule=0pt,
  arc=0pt,
  outer arc=0pt,
  colframe=orange,
}

\newtcolorbox{myboxii}[1][]{
  breakable,
  freelance,
  title=#1,
  colback=white,
  colbacktitle=white,
  coltitle=black,
  fonttitle=\bfseries,
  bottomrule=0pt,
  boxrule=0pt,
  colframe=white,
  overlay unbroken and first={
  \draw[red!75!black,line width=3pt]
    ([xshift=5pt]frame.north west) -- 
    (frame.north west) -- 
    (frame.south west);
  \draw[red!75!black,line width=3pt]
    ([xshift=-5pt]frame.north east) -- 
    (frame.north east) -- 
    (frame.south east);
  },
  overlay unbroken app={
  \draw[red!75!black,line width=3pt,line cap=rect]
    (frame.south west) -- 
    ([xshift=5pt]frame.south west);
  \draw[red!75!black,line width=3pt,line cap=rect]
    (frame.south east) -- 
    ([xshift=-5pt]frame.south east);
  },
  overlay middle and last={
  \draw[red!75!black,line width=3pt]
    (frame.north west) -- 
    (frame.south west);
  \draw[red!75!black,line width=3pt]
    (frame.north east) -- 
    (frame.south east);
  },
  overlay last app={
  \draw[red!75!black,line width=3pt,line cap=rect]
    (frame.south west) --
    ([xshift=5pt]frame.south west);
  \draw[red!75!black,line width=3pt,line cap=rect]
    (frame.south east) --
    ([xshift=-5pt]frame.south east);
  },
}

\tcbset{
  takeawaysbox/.style={
    title=Takeaways,
    colback=lightblue!80,
    colframe=black,
    fonttitle=\bfseries\small,
    coltitle=white,
    colbacktitle=black,
    enhanced,
    attach boxed title to top left={xshift=2.5mm,yshift=-2.5mm},
    boxed title style={rounded corners, size=small, colframe=black, colback=black},
    width=\linewidth,
    arc=3.5mm
  }
}

\mdfdefinestyle{mystyle}{%
  rightline=true,
  innerleftmargin=10,
  innerrightmargin=10,
  outerlinewidth=3pt,
  topline=false,
  rightline=true,
  bottomline=false,
  skipabove=\topsep,
  skipbelow=\topsep
}

\tikzset{%
    every node/.style={font=\tiny},
    parent/.style =          {align=center,text width=2cm,rounded corners=3pt, line width=0.3mm, fill=gray!10,draw=gray!80},
    child/.style =           {align=center,text width=2.0cm,rounded corners=3pt, fill=blue!10,draw=blue!80,line width=0.3mm},
    grandchild/.style =      {align=center,text width=2cm,rounded corners=3pt},
    greatgrandchild/.style = {align=center,text width=1.5cm,rounded corners=3pt},
    greatgrandchild2/.style = {align=center,text width=1.5cm,rounded corners=3pt},    
    referenceblock/.style =  {align=center,text width=1.5cm,rounded corners=2pt},
    pretrain/.style =           {align=center,text width=2.0cm,rounded corners=3pt, fill=blue!10,draw=blue!80,line width=0.3mm},   
    pretrain_work/.style =           {align=center, text width=8.5cm,rounded corners=3pt, fill=blue!10,draw=blue!0,line width=0.3mm},  
    template/.style =           {align=center,text width=2.0cm,rounded corners=3pt, fill=red!10,draw=red!80,line width=0.3mm},   
    template_work/.style =           {align=center,text width=8.5cm,rounded corners=3pt, fill=red!10,draw=red!0,line width=0.3mm},    
    answer/.style =           {align=center,text width=2.0cm,rounded corners=3pt, fill= cyan!10,draw= cyan!80,line width=0.3mm},   
    answer_work/.style =           {align=center,text width=8.5cm,rounded corners=3pt, fill= cyan!10,draw= cyan!0,line width=0.3mm},      
    multiple/.style =           {align=center,text width=2.0cm,rounded corners=3pt, fill= orange!10,draw= orange!80,line width=0.3mm},   
    multiple_work/.style =           {align=center,text width=8.5cm,rounded corners=3pt, fill= orange!10,draw= orange!0,line width=0.3mm},        
    tuning/.style =           {align=center,text width=2.0cm,rounded corners=3pt, fill= magenta!10,draw= magenta!80,line width=0.3mm},   
    tuning_work/.style =           {align=center,text width=8.5cm,rounded corners=3pt, fill= magenta!10,draw= magenta!0,line width=0.3mm},          
}

\newcommand{\lstbg}[3][0pt]{{\fboxsep#1\colorbox{#2}{\strut #3}}}

\lstdefinelanguage{diff}{
  basicstyle=\ttfamily\small,
  morecomment=[f][\lstbg{red!20}]-,
  morecomment=[f][\lstbg{green!20}]+,
}

\lstdefinelanguage{diffpython}{
  language=diff,
  morekeywords={def, if, else, for, while, return, import, from, as, class, with, try, except, finally, raise, lambda, and, or, not, in, is, None, True, False},
  morecomment=[l]{\#},
  morestring=[b]",
  morestring=[b]',
}

\setheadertext{LUMIA Lab}

\correspondingemail{\emailicon\ hollyming@sjtu.edu.cn, lin.zhouhan@gmail.com \quad $^\ddagger$ Corresponding Author. }

\githublink{https://github.com/LUMIA-Group/habit-bench}

\huggingfacelink{https://huggingface.co/datasets/hollyming/HABIT-Bench}

\renewcommand{\today}{2026-10-17}

\definecolor{habitCI}{HTML}{737373}
\definecolor{habitGroupBlue}{HTML}{E8F1FA}
\definecolor{habitAverageGray}{HTML}{F0F1F3}
\definecolor{uclablue}{RGB}{39, 116, 174}
\definecolor{bigaired}{RGB}{156, 0, 0}

\title{HABIT-Bench: Evaluating Habit Induction from Longitudinal Weak Evidence in Agent Memory}
\setheadertitle{HABIT-Bench}

\author{%
  Junming Zhang$^{1,2}$, Rubin Wei$^{1,2}$, Xu Cheng$^{1,2}$, Hao Sun$^{1,2}$, Jiarui Wang$^{1,2}$, Ruoxi Cheng$^{3}$, Xiquan Wang$^{1}$, Peisong Wu$^{1}$, Mingsong Chen$^{1}$,\
  \Authfont Zhouhan Lin$^{1,2\ddagger}$\\
  {\Affilfont $^1$ LUMIA Lab, School of Artificial Intelligence, Shanghai Jiao Tong University} \\
  {\Affilfont $^2$ Shanghai Artificial Intelligence Laboratory} \quad
  {\Affilfont $^3$ Fudan University}
}

\begin{document}

\begin{abstract}
In long-term interactions with large language model (LLM) agents, users reveal habits through their choices and actions.
These habits capture implicit user information that agents must infer across conversations.
However, individual observations may provide only partial evidence about a habit or when it applies. 
Retaining past interactions alone does not ensure that agents can infer habits and apply them appropriately. 
We introduce HABIT-Bench to evaluate how well agent memory systems infer habits from weak evidence and apply them to later requests. 
We construct synthetic user histories from task-oriented dialogue data and controlled habit graphs. 
The histories span hundreds of sessions across food, finance, software, and travel. 
Four-choice tasks assess agents' ability to infer user habits and apply them appropriately. 
They include temporary exceptions, changes in habits, and suggestions that users have not accepted. 
Sampled tasks undergo independent human review. 
We evaluate eight memory methods with Qwen3-235B-A22B and GLM-5.3-Flash. 
The best-performing memory method achieves 33.61\% macro accuracy across the four domains.
Performance remains limited even when annotated evidence is supplied directly. These results expose a persistent gap between access to past interactions and the ability to infer and appropriately use user habits.

\end{abstract}

\maketitle


\input{sec/1_introduction}
\input{sec/2_related_work}
\input{sec/3_habit_bench}
\input{sec/4_experiments}
\input{sec/5_conclusion}

\bibliography{references}

\clearpage
\appendix
\section*{Appendix}
\input{sec/6_appendix}

\end{document}

%% file: notations.tex
\usepackage{xspace}

\theoremstyle{plain}

\newtheorem*{proposition*}{Proposition}

\theoremstyle{definition}

\theoremstyle{definition}

\newcommand{\A}{\mathcal{A}}

\def\eqref#1{equation~\ref{#1}}



%% file: sec/1_introduction.tex
\section{Introduction}
Long-term interaction requires large language model (LLM) agents to learn from experience, not merely retain explicit facts and preferences \citep{shinn2023reflexion,xu2026mem}.
Associative and latent memories offer complementary mechanisms for organizing and retaining experience \citep{gutierrez2024hipporag,wang2024memoryllm,wang2025m+}.
Useful personalization depends on inferring user characteristics and preferences from interaction history \citep{zhong2024memorybank,zollo2025personalllm,nam2026learning} and deciding when these regularities should guide a response.
Context and subsequent changes can alter whether an otherwise useful default applies \citep{wood2016psychology,jiang2025know}.

\input{tables/benchmark_comparison}

Existing benchmarks address complementary parts of this problem.
LoCoMo and LongMemEval assess recall and reasoning over extended histories \citep{maharana2024evaluating,wu2024longmemeval}, while MemoryAgentBench adds incremental learning and selective forgetting \citep{hu2026evaluating}.
PrefEval examines implicit preference inference and following, while PersonaMem tracks evolving user profiles \citep{zhao2025llms,jiang2025know}.
PERMA and DynamicMem study evolving preferences and habits \citep{liu2026perma,xie2026dynamicmem}.
BenchPreS and OP-Bench examine when personalization should be withheld \citep{yoon2026benchpres,hu2026op}, while MemoryArena and Mem2ActBench connect memory to subsequent action \citep{he2026memoryarena,shen2026mem2actbench}.
These developments make a precise evaluation question important: when does the available history justify applying a user-specific behavioral default to a new decision?


We introduce HABIT-Bench to study this question. We define habit memory as the ability to infer behavioral defaults from incomplete evidence across sessions and apply them according to their scope, exceptions, temporal validity, and user endorsement. 
A relevant past interaction need not establish a valid default. A repeated choice may apply only in one context, and an assistant's suggestion does not become a user preference merely because it appears in the history. The target is the support that memory provides for a justified decision, rather than a particular memory representation \citep{rabinowitz2018machine,siththaranjan2024distributional}.


HABIT-Bench constructs synthetic histories using task-oriented dialogue seeds and controlled habit graphs. 
It contains 114 users, 37,152 sessions, and 3,328 probes across food, finance, software, and travel.
The domains cover complementary evidence structures. Food combines repeated choices with contextual contrasts, and Travel tests behavioral-rule recovery and its application to new requests. Finance and Software emphasize reconstructing user-confirmed workflows and tracking their changes across sessions. Candidate actions differ in their use of habits, including applying an outdated choice, extending a default beyond its scope, or treating an unaccepted suggestion as an established decision.


Memory methods process each user's history chronologically and supply context to a shared base model. We evaluate eight memory methods alongside retrieval baselines and a compressed-history control. The best memory-method macro accuracy is 33.06\% with Qwen3-235B-A22B \citep{yang2025qwen3} and 33.61\% with GLM-5.3-Flash \citep{zeng2026glm}. These results are close to the best retrieval baselines, while GLM's compressed-history control reaches 41.98\%. Oracle Evidence improves access to relevant observations but still leaves substantial decision errors. The results identify both the reconstruction of habits and their appropriate use as important targets for memory system evaluation.


Our contributions are as follows.

\textbf{Capability definition.} 
We formulate habit memory as the joint problem of inferring behavioral defaults and determining when the evidence supports their use.

\noindent\textbf{Benchmark construction.} 
We introduce a four-domain benchmark with chronological histories, annotated habit states and supporting sessions, and human validation of decision recoverability.

\textbf{Evaluation and diagnostics.} 
We compare memory systems under a shared answer protocol and provide diagnostic controls for evidence access and downstream decision making. The experiments show that elaborate memory pipelines do not consistently outperform simple retrieval on these tasks.



%% file: tables/benchmark_comparison.tex
\begin{table*}[t]
\centering
\definecolor{benchmarkGreen}{HTML}{207A55}
\definecolor{benchmarkRed}{HTML}{B83C4A}
\newcommand{\benchfull}{\textcolor{benchmarkGreen}{\ensuremath{\boldsymbol{\checkmark}}}}
\newcommand{\benchpartial}{\ensuremath{\circ}}
\newcommand{\benchnone}{\textcolor{benchmarkRed}{\ensuremath{\boldsymbol{\times}}}}
\caption{Capability coverage across representative long-term-memory and personalization benchmarks. \benchfull, \benchpartial, and \benchnone denote dedicated, partial, and absent benchmark-level coverage, respectively. Columns cover longitudinal history, weak-evidence induction, memory-guided action, contextual gating, exceptions, temporal drift, and withholding unsupported personalization.}
\label{tab:benchmark-comparison}
\setlength{\tabcolsep}{4.0pt}
\renewcommand{\arraystretch}{1.12}
\scriptsize
\begin{tabular*}{\textwidth}{@{\extracolsep{\fill}}l*{7}{c}@{}}
\toprule
\textbf{Benchmark} & \textbf{Long Hist.} & \textbf{Weak Evid.} & \textbf{Action} & \textbf{Context Gate} & \textbf{Exception} & \textbf{Temp. Drift} & \textbf{Withhold} \\
\midrule
LoCoMo \citep{maharana2024evaluating} & \benchfull & \benchnone & \benchnone & \benchnone & \benchnone & \benchpartial & \benchnone \\
LongMemEval \citep{wu2024longmemeval} & \benchfull & \benchnone & \benchnone & \benchnone & \benchnone & \benchfull & \benchnone \\
MemoryAgentBench \citep{hu2026evaluating} & \benchfull & \benchpartial & \benchpartial & \benchnone & \benchnone & \benchfull & \benchnone \\
\addlinespace[1.5pt]
MemoryArena \citep{he2026memoryarena} & \benchfull & \benchnone & \benchfull & \benchpartial & \benchnone & \benchnone & \benchnone \\
Mem2ActBench \citep{shen2026mem2actbench} & \benchfull & \benchnone & \benchfull & \benchpartial & \benchnone & \benchfull & \benchnone \\
\addlinespace[1.5pt]
PERMA \citep{liu2026perma} & \benchfull & \benchfull & \benchfull & \benchpartial & \benchpartial & \benchfull & \benchnone \\
DynamicMem \citep{xie2026dynamicmem} & \benchfull & \benchfull & \benchfull & \benchpartial & \benchpartial & \benchfull & \benchnone \\
HorizonBench \citep{li2026horizonbench} & \benchfull & \benchpartial & \benchnone & \benchnone & \benchnone & \benchfull & \benchnone \\
IBA-Bench \citep{song2026profiling} & \benchfull & \benchfull & \benchfull & \benchpartial & \benchpartial & \benchfull & \benchpartial \\
\addlinespace[1.5pt]
BenchPreS \citep{yoon2026benchpres} & \benchpartial & \benchnone & \benchfull & \benchfull & \benchnone & \benchnone & \benchfull \\
OP-Bench \citep{hu2026op} & \benchfull & \benchnone & \benchfull & \benchfull & \benchnone & \benchnone & \benchfull \\
MemOps \citep{hao2026memops} & \benchfull & \benchnone & \benchpartial & \benchpartial & \benchpartial & \benchfull & \benchpartial \\
\addlinespace[1.5pt]
PersonaTrail \citep{yang2026personatrail} & \benchfull & \benchfull & \benchfull & \benchpartial & \benchnone & \benchnone & \benchnone \\
PersonalAlign \citep{lyu2026personalalign} & \benchfull & \benchfull & \benchfull & \benchfull & \benchpartial & \benchpartial & \benchpartial \\
RefMem-Bench \citep{lin2026connecting} & \benchfull & \benchfull & \benchpartial & \benchpartial & \benchnone & \benchfull & \benchpartial \\
\midrule
\textbf{HABIT-Bench} & \benchfull & \benchfull & \benchfull & \benchfull & \benchfull & \benchfull & \benchfull \\
\bottomrule
\end{tabular*}
\vspace{-8pt}
\end{table*}

%% file: sec/2_related_work.tex
\section{Related Work}
\textbf{Long-term memory evaluation.}
LoCoMo evaluates long-conversation question answering, summarization, and multimodal generation \citep{maharana2024evaluating}, while LongMemEval tests extraction, multi-session and temporal reasoning, updates, and abstention \citep{wu2024longmemeval}.
MemoryAgentBench examines retrieval, learning, long-range understanding, and selective forgetting under incremental input \citep{hu2026evaluating}, and MemOps diagnoses memory operations and their composition \citep{hao2026memops}.

\textbf{Implicit and evolving user modeling.}
PERMA studies preference formation and refinement \citep{liu2026perma}, while DynamicMem evaluates evolving attributes, habits, and preferences across applications \citep{xie2026dynamicmem}.
HorizonBench constructs graph-controlled preference trajectories with traceable provenance \citep{li2026horizonbench}.
RefMem-Bench tests behavioral pattern induction, temporal dynamics, and implicit boundaries from fragmented multimodal evidence \citep{lin2026connecting}.
HABIT-Bench focuses on whether longitudinal evidence supports a behavioral default and its subsequent application.

\textbf{Selective personalization.}
BenchPreS tests preference application under social and institutional constraints in third-party communication \citep{yoon2026benchpres}.
OP-Bench examines irrelevant memory use, repetition, and sycophancy \citep{hu2026op}.
HABIT-Bench couples selectivity with weak-evidence induction, testing scope, exceptions, temporal validity, and user endorsement.

\textbf{Memory-grounded agent behavior.}
MemoryArena evaluates experience reuse across interdependent sessions \citep{he2026memoryarena}, and Mem2ActBench tests memory-dependent tool selection and parameter grounding \citep{shen2026mem2actbench}.
PersonaTrail studies personalized web navigation \citep{yang2026personatrail}, while PersonalAlign's AndroidIntent evaluates intent resolution and proactive GUI assistance \citep{lyu2026personalalign}.
IBA-Bench connects implicit user modeling to task execution under enduring traits and transient states \citep{song2026profiling}.
HABIT-Bench uses non-interactive action selection with a shared answer LLM to compare the context produced by different memory systems. 
Table~\ref{tab:benchmark-comparison} compares these evaluation targets.


%% file: sec/3_habit_bench.tex
\section{HABIT-Bench}
\label{sec:habitbench}
\newcommand{\habitz}{z}
\newcommand{\ind}{\mathbb{I}}

\begin{figure}[tb]
\centering
\includegraphics[width=\linewidth]{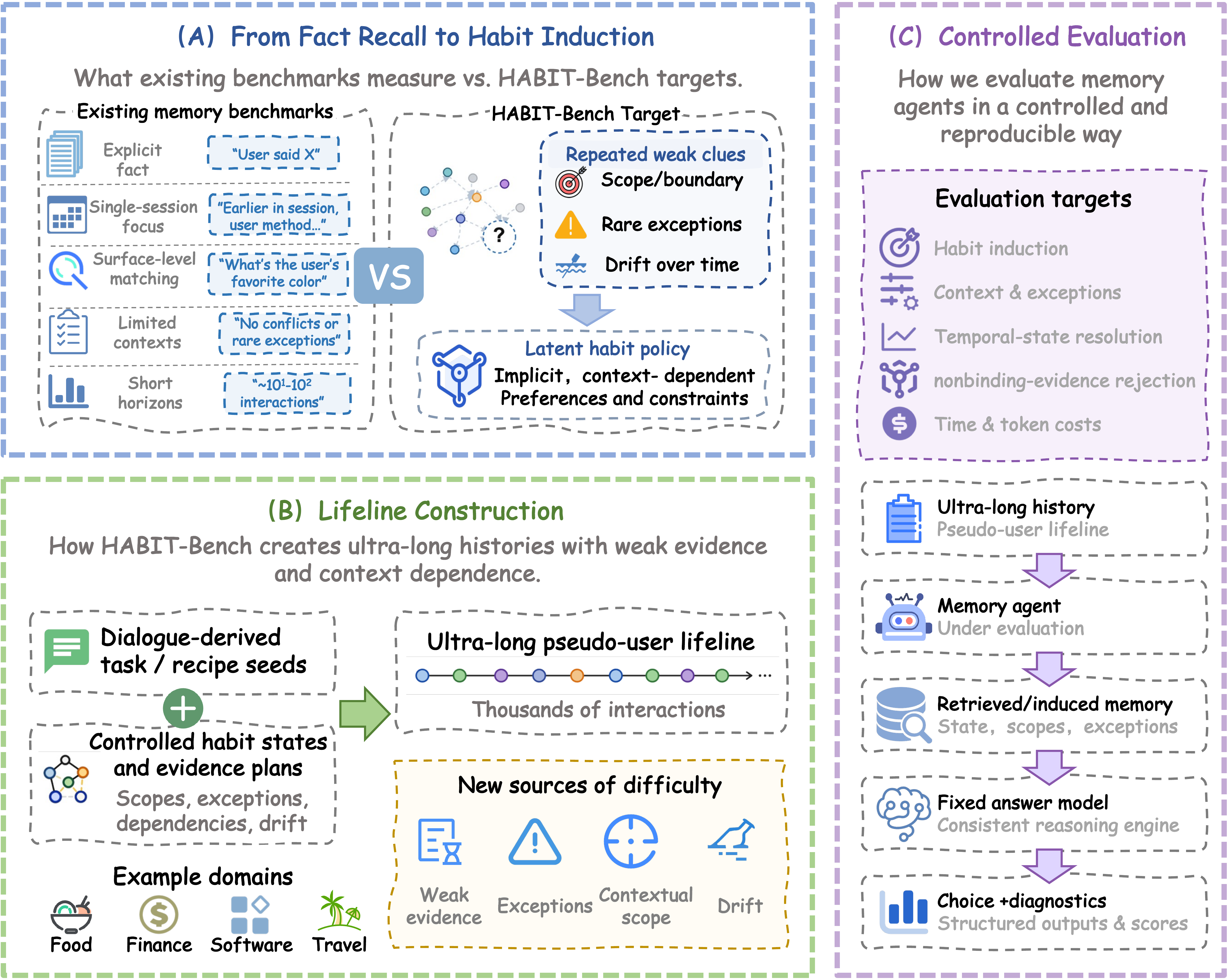}
\caption{HABIT-Bench framework. (A) contrasts explicit-memory retrieval with latent habit induction. (B) combines task-oriented prompt seeds with latent habit graphs to construct real-prompt-seeded synthetic lifelines. (C) evaluates memory systems through a shared answerer and diagnostic choice interface.}
\label{fig:framework}
\end{figure}

HABIT-Bench uses four-choice tasks to evaluate whether agents can infer user habits and use them appropriately to select actions. 
Figure~\ref{fig:framework} outlines the benchmark, and Appendix~\ref{app:case-studies} illustrates how distributed evidence determines the valid action in each domain.

\subsection{Task Formulation}
\label{sec:task}
For user $u$, let $H_u^{\le t}=(s_1,\ldots,s_t)$ denote the history through session $t$.
A probe $p$ contains a request $x_p$, four candidate actions $A_p$, a visibility cutoff $t_p$, and a target time $\tau_p$ specifying the current or historical state to resolve.
Each habit state $z_{u,h}(\tau)$ describes a behavioral default, its applicability, exceptions, temporal validity, and user endorsement. 
The evaluator specifies
\begin{equation}
\begin{aligned}
Z_u(\tau)&=\{\habitz_{u,h}(\tau):h\in\mathcal B_u\},  \qquad
a_p^\star&=\operatorname{Resolve}\!\left(x_p,A_p,Z_u(\tau_p)\right),
\end{aligned}
\label{eq:habit-decision}
\end{equation}
where $\mathcal B_u$ indexes the user's habits. Resolve combines relevant habits and enforces the request's constraints to identify one valid action. This specification leaves the evaluated method's memory representation unrestricted.

We define weak evidence as observations that provide incomplete support for a default or its validity conditions. A recurring choice can leave scope uncertain, and a later selection can require an earlier shortlist for interpretation. 
The task therefore combines behavioral inference and cross-session decision reconstruction. In the Finance example (Appendix Figure~\ref{fig:case-finance}), two workflows must be reconstructed at a May 2023 target time even though their 2025 replacements are visible. Correct memory use requires preserving both the chosen actions and when they became effective.


\subsection{Data and Task Construction}
\label{sec:data}


HABIT-Bench contains 114 synthetic users, 37,152 sessions, and 3,328 probes across four domains (Table~\ref{tab:datasets}). 
Source datasets provide task seeds and interaction patterns. HABIT-Bench constructs user continuity, habit states, and longitudinal evidence relationships. Histories interleave habit-relevant observations with ordinary same-domain tasks, while private annotations link probes to applicable states and supporting sessions. The four domains instantiate complementary evidence structures.

\input{tables/dataset_overview.tex}


\textbf{Food.}
We use cooking dialogues and recipes from CookDial~\citep{jiang2023cookdial} to create conversational evidence of preparation and serving habits. 
Earlier alternatives, later user selections, and repeated use establish behavioral defaults. 
Contrasting contexts and local exceptions determine their scope. We rewrite the sessions as natural conversations while preserving recipe facts and habit evidence. Direct-use, boundary, and exception probes distinguish applying a default from overgeneralizing it.


\textbf{Finance.}
We use banking dialogues from MultiDoGO~\citep{peskov2019multi} to construct histories of financial tasks.
Ordered shortlists and later user confirmations establish or replace standing choices. Two- and three-habit probes combine workflows at current or historical times, with local exceptions, unresolved cases, and unaccepted suggestions providing competing evidence.


\textbf{Software.}
The Software histories use technical-support dialogues from MultiDoGO~\citep{peskov2019multi}. Each user has a consistent software environment. We use the same selection process as Finance, with more frequent changes to the chosen procedures. The histories also include similar errors in different environments and unfinished support cases. Questions combine two or three habits and may require continuing an earlier case.


\textbf{Travel.}
Taskmaster-2~\citep{byrne2019taskmaster} flight and hotel dialogues supply behavioral candidates for synthetic user profiles. Chronological histories are authored before evidence annotation and blind recoverability checks. Questions involve one or two habits and test their application to new requests, including exceptions and changes over time. 


\textbf{Choices and validation.}
Alternatives are designed to meet the request's hard constraints while differing in habit use and validity. 
We check that the choices cover the request, the evidence supports the intended decision, and the context is consistent. Domain-specific structural checks, model-based checks, and human review also assess whether a choice can be identified without using the history. 
Appendix~\ref{app:domain-construction} details templates, evidence schedules, source transformations, and validation. The resulting lifelines are source-grounded synthetic histories, not longitudinal records of the source participants.



\subsection{Evaluation Protocol and Metrics}
\label{sec:protocol}
Methods ingest sessions chronologically into user-isolated memory, $M_u^t=\operatorname{Update}(M_u^{t-1},s_t)$. At probe $p$, they construct $R_p=\operatorname{Retrieve}(M_u^{t_p},x_p)$ under recorded method-specific budgets. A shared answerer within each comparison predicts $\hat a_p=\operatorname{Answer}(x_p,\mathcal{A}_p,R_p)$. Standard methods cannot access private gold choices, habit states, evidence annotations, future sessions, or other users' memories. 
Appendix~\ref{app:implementation-provenance} documents their adapters.

For the evaluated probe set $P_d$ in domain $d$, the primary scores are
\begin{equation}
\operatorname{Acc}_d=
\frac{1}{|P_d|}
\sum_{p\in P_d}\mathbf{\ind}[\hat a_p=a_p^\star],
\qquad
\operatorname{Macro}=
\frac{1}{4}\sum_{d=1}^{4}\operatorname{Acc}_d.
\end{equation}

Exact-match scoring of the four-choice predictions avoids variability introduced by LLM judges and does not require a separate, stronger judge model.
Uniform random choice yields 25\%. Appendix~\ref{app:model-scaling} reports supplementary user-macro scores and available 95\% user-cluster bootstrap intervals.
Evidence-access and answer-evidence diagnostics require source-session attribution and eligible annotations. Missing attribution is not counted as retrieval failure.
Appendices~\ref{app:capability-slices} and~\ref{app:retrieval-diagnostics} report capability slices and evidence metrics, respectively. Section~\ref{subsec:retreval_sensitivity_and_memory_costs} reports build/query time and LLM token usage. Boundary and nonbinding-evidence diagnostics do not constitute probability calibration or a global false-personalization rate.

\subsection{Human Validation}
\label{sec:human-validation}

Human validation tests whether the intended decision is recoverable from the annotated evidence, independently of a memory system's ability to find that evidence in a long history.
Two human reviewers, A and B, independently inspect the request, four alternatives, and a chronological evidence packet without seeing the gold answer or system predictions.
They select the uniquely supported action and assess eight criteria covering answerability, evidence sufficiency, scope, boundaries and exceptions, choice balance, naturalness, source grounding, and privacy.
An item without a uniquely defensible answer receives no choice rather than a forced guess.
A third human reviewer, C, adjudicates disagreements in either the answer or quality judgments before the merged record is frozen and compared with the gold labels.
Appendix~\ref{app:human-audit} details the protocol, sampling, and domain-specific results.



Independent review covers 200 Food, 340 Finance, 311 Software, and 618 Travel probes.
A/B answer agreement over jointly valid choices is 99.50\%, 100.00\%, 99.68\%, and 92.84\%, respectively, with domain-specific denominators and Cohen's $\kappa$ reported in Appendix Table~\ref{tab:human-audit-agreement}.
The quality rubric is stricter than agreement with gold: four Finance items initially receive Modify because their alternatives omit an additional output-placement requirement, although the evidence supports the intended habit.
These local issues were corrected before final evaluation, while the audit retains the original modification flags.
The remaining 336 Finance items and all 311 Software items pass adjudication.
All 79 content-revised Travel items pass targeted gold-blind validation, and all 200 reviewed Food items pass naturalness checks after naturalization.
High independent agreement and the rubric's ability to detect narrow construction defects support the quality of the reviewed evidence and decision labels.
These checks assess decision recoverability given the evidence, not human retrieval from full histories.

%% file: tables/dataset_overview.tex
\begin{table}[t]
\caption{Dataset statistics and source datasets. Sources provide dialogue and task seeds, while HABIT-Bench constructs the longitudinal histories and probe labels.}
\label{tab:datasets}
\centering
\small
\begin{tabular}{lrrrl}
\toprule
Domain & Users & Sessions & Probes & Source dataset \\
\midrule
Food & 30 & 3900 & 630 & CookDial~\citep{jiang2023cookdial} \\
Finance & 36 & 19440 & 1368 & MultiDoGO~\citep{peskov2019multi} \\
Software & 18 & 9720 & 680 & MultiDoGO~\citep{peskov2019multi} \\
Travel & 30 & 4092 & 650 & Taskmaster-2~\citep{byrne2019taskmaster} \\
\bottomrule
\end{tabular}
\end{table}

%% file: sec/4_experiments.tex
\section{Experiments}
\label{sec:results}

\subsection{Experimental Setup}
\textbf{Systems and models.}
We organize the comparison by the role of memory in each system: controls, retrieval baselines, memory methods, and multi-store memory methods.
The No Memory and Full Memory controls omit the interaction history or provide a compressed history, respectively.
The retrieval baselines select sessions directly from the history and comprise Recency-5, Recency-10, BM25 RAG, Dense RAG, and Temporal Hybrid RAG.
The memory methods construct and maintain their own memory representations:  A-MEM~\citep{xu2026mem}, Mem0~\citep{chhikara2025mem0}, MemOS~\citep{li2025memos}, MemRL~\citep{zhang2026memrl}, LightMem~\citep{fang2026lightmem}, and SeCom~\citep{pan2025memory}. 
Letta~\citep{packer2023memgpt} and MIRIX~\citep{wang2025mirix} are shown separately as multi-store memory methods, reflecting their coordination of distinct memory stores.
The other six are grouped for presentation rather than treated as a single architecture.
References to the eight memory methods include both groups.
Qwen3-235B-A22B~\citep{yang2025qwen3} and GLM-5.3-Flash~\citep{zeng2026glm} are the two primary language models.
Within each setting, all methods use the same model to select an answer and to perform memory operations that require an LLM.
Dense retrieval components use a fixed BGE-M3 encoder~\citep{chen2025m3embedding}.
Together with additional controls, oracle diagnostics, and experiments with smaller Qwen3 models, the evaluation covers 308 configurations across methods, models, and domains.


\textbf{Inference protocol.}
Methods maintain a separate memory for each user and process sessions in chronological order, using only the history available at each probe's visibility cutoff.
We set the decoding temperature to zero.
The two primary model settings use the same prompt for selecting an action and require outputs to follow a fixed JSON schema.
In the GLM experiments, the Qwen3-8B tokenizer is used only to measure token budgets, not to generate memories or answers.
Context construction and token budgets vary by method.
We report accuracy over all probes within each domain and macro accuracy as the unweighted mean of the four domain accuracies.
Appendix~\ref{app:model-scaling} reports supplementary accuracy averaged across users and, where available, bootstrap confidence intervals obtained by resampling users.
It also provides the model configurations, decision prompts, token budgets, and rules for truncation, retries, and answer parsing.


\textbf{Baselines and oracle controls.}
Retrieval baselines use the request alone, without the answer alternatives, to select complete sessions from the visible history.
No Memory omits the interaction history, while Full Memory supplies a compacted history that fits within its context budget.
Oracle Evidence supplies the private annotated evidence, whereas Oracle Habit State supplies the private habit state and required action.
Both oracle conditions use information unavailable to the evaluated methods and serve only as diagnostic controls.

\subsection{Main Results}

\input{tables/dual_model_results.tex}


Table~\ref{tab:main-results} evaluates whether current memory systems can infer habits from longitudinal weak evidence and apply them to subsequent decisions. Three findings emerge across the two primary model settings.

\textbf{Habit induction and action selection remain challenging.}
Across the eight memory methods, four-domain macro accuracy ranges from 27.01\% to 33.06\% with Qwen3-235B-A22B and from 30.42\% to 33.61\% with GLM-5.3-Flash.
Even the best configurations achieve only approximately one-third accuracy, despite using the primary model for both memory operations and action selection.
The task requires selecting one of four actions and is scored by exact match, without an LLM judge.
The model need not formulate a free-form response, but must still distinguish actions supported by the user's history from plausible yet unsupported alternatives. 
Habit-conditioned decision making remains a demanding evaluation target across the tested memory designs and model settings.

\textbf{The benefit of memory depends on the domain.}
Under Qwen3-235B-A22B, the best memory-method results improve over No Memory by 20.32 percentage points on Food and 13.07 points on Travel.
In Finance, all eight methods fall below the 23.54\% No Memory point estimate. In Software, the largest gain is 1.03 percentage points.
The domain contrast also appears with GLM: the best memory-method accuracies are 28.29\% on Finance and 25.44\% on Software, compared with 54.62\% on Travel.
Finance and Software require agents to combine multiple habits, resolve current or historical choices, and distinguish user commitments from unaccepted suggestions.
Their low scores show that the evaluated methods struggle with these combined requirements, but domain comparisons alone cannot identify which requirement causes an error.
Memory can substantially improve decisions in some domains without providing reliable benefits in others.
Section~\ref{subsec:diagnostic_controls_and_evidence_analysis} examines this distinction through evidence and capability diagnostics.

\textbf{Memory methods do not consistently outperform retrieval.} 
The best memory-method macro accuracy is close to that of simple retrieval: 33.06\% versus 32.50\% for BM25 with Qwen, and 33.61\% versus 33.21\% for dense retrieval with GLM.
GLM's Full Memory control reaches 41.98\%, exceeding all eight memory methods while leaving substantial room for improvement.
Extracting and organizing memories therefore does not consistently produce better decisions than retrieving sessions or compressing history in these experiments.
The methods differ in context construction and effective budgets, so these results compare complete pipelines rather than isolate the effect of a memory architecture.
The relevant question is whether a representation preserves the evidence and conditions needed for a later decision, not simply whether it stores a compact account of the history.

\subsection{Diagnostic Controls and Evidence Analysis}
\label{subsec:diagnostic_controls_and_evidence_analysis}

\begingroup
\setlength{\tabcolsep}{5pt}
\input{tables/diagnostic_results.tex}
\endgroup

Annotated evidence leaves substantial decision difficulty. Under Qwen3-235B-A22B, Oracle Evidence achieves 39.26–58.25\% accuracy across domains, while Oracle Habit State achieves 98.41–99.26\% (Table ~\ref{tab:diagnostic-results}). The same qualitative separation appears with Qwen3-8B. Supplying the annotated evidence therefore does not by itself resolve the task. Oracle Habit State additionally exposes the private active state and required action, demonstrating that the answerer can map this privileged information to the correct alternative. The comparison diagnoses remaining difficulty after evidence selection, while combining differences in reconstruction requirements and supplied information.


Evidence access and decision correctness remain distinct. For the six source-attributed Qwen3-32B methods, Hit@5 ranges from 37.77\% to 41.44\%, and annotated-evidence recall from 11.17\% to 13.89\% (Appendix~\ref{app:retrieval-diagnostics} Table~\ref{tab:qwen32b-retrieval}). 
Hit@5 indicates whether any of the first five attributed sessions is annotated as positive evidence. Evidence recall measures the fraction of the annotated set covered by these sessions. 
The joint accuracy of the answer and the evidence (Joint A+E) is 12.92-16.20\%, requiring a correct answer and at least one evidence hit. Clean Grounded@5 is 5.49-8.20\% on its eligible subset and additionally requires that none of the attributed sessions be labeled nonbinding. 
This stricter condition describes the composition of the attributed context. It does not directly measure whether the answerer improperly follows a retrieved suggestion. 
These diagnostics distinguish evidence access, answer correctness, and the exclusion of nonbinding observations, without interpreting an evidence hit as complete reasoning support or differences between eligible subsets as an error decomposition.

This distinction matters when the decisive information is distributed across sessions.
Retrieving a user's later selection may be insufficient if an earlier session defines the alternatives, just as retrieving an established default may be insufficient without a subsequent exception or replacement.
An evidence hit can therefore identify a relevant observation without establishing which action is justified at the probe's target time.
The diagnostic goal is to assess whether the retrieved context supports the complete decision, including the conditions under which the remembered behavior applies.

The capability slices in Appendix~\ref{app:capability-slices} provide a complementary analysis.
Method rankings vary across temporal resolution, provenance, reference reconstruction, induction, and exception handling.
Strong relative performance on one requirement does not imply the same ordering on another.
Overall accuracy should thus be read alongside these diagnostics, which identify distinct demands on memory without attributing every error to retrieval or to habit induction alone.

\subsection{Qwen Model-Size Comparison}

\begin{figure}[ht]
\centering
\includegraphics[width=\linewidth]{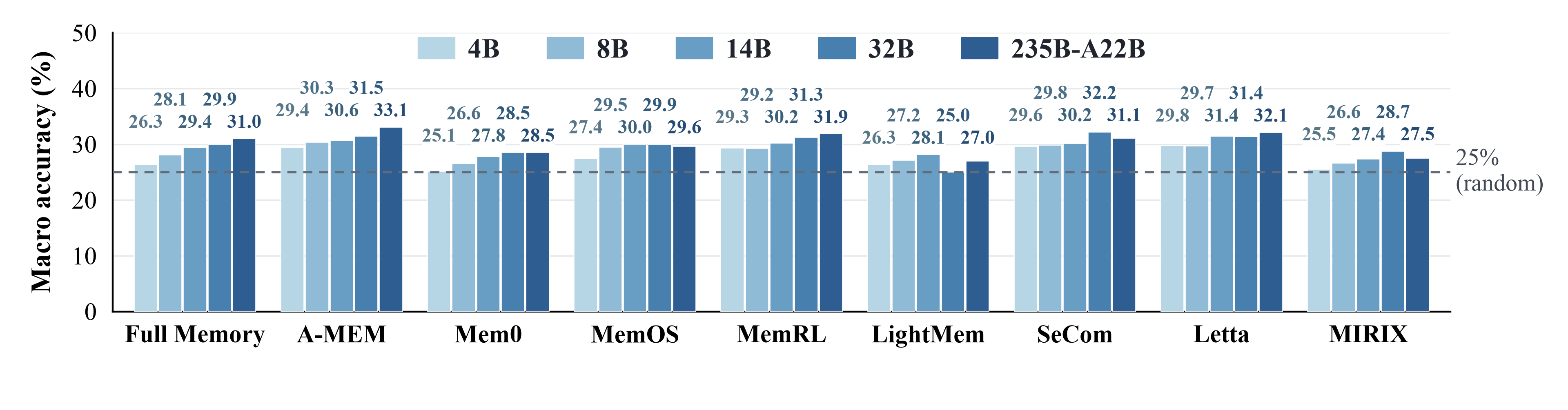}
\vspace{-24pt}
\caption{Qwen model-size comparison. Each group shows four-domain macro accuracy for five model settings. Full Memory is a control, and the dashed line marks 25\% uniform random choice. Both memory-stage and answer models change, and the 235B setting uses a revised reader protocol, so the comparison does not isolate parameter scale.}
\label{fig:qwen-scaling}
\end{figure}

Figure~\ref{fig:qwen-scaling} compares Qwen3-4B, 8B, 14B, 32B, and 235B-A22B using the same four-domain mean of probe-level accuracy.
The exact values are retained in Appendix Table~\ref{tab:qwen-scale-micro}.
Larger models do not yield uniform gains across methods.
A-MEM increases from 29.41\% with 4B to 33.06\% with 235B, whereas LightMem reaches 28.14\% with 14B, drops to 25.04\% with 32B, and recovers to 27.01\% with 235B.
These comparisons do not isolate a scale-only effect because both memory-stage and answer models change, and the smaller Qwen runs use an earlier reader protocol.
Nevertheless, the modest and non-monotonic differences show that the larger model settings do not consistently resolve the benchmark's difficulties.

\subsection{Retrieval Sensitivity and Memory Costs}
\label{subsec:retreval_sensitivity_and_memory_costs}

\begin{figure}[htbp]
\centering
\begin{minipage}[t]{0.485\linewidth}
\centering
\includegraphics[width=\linewidth]{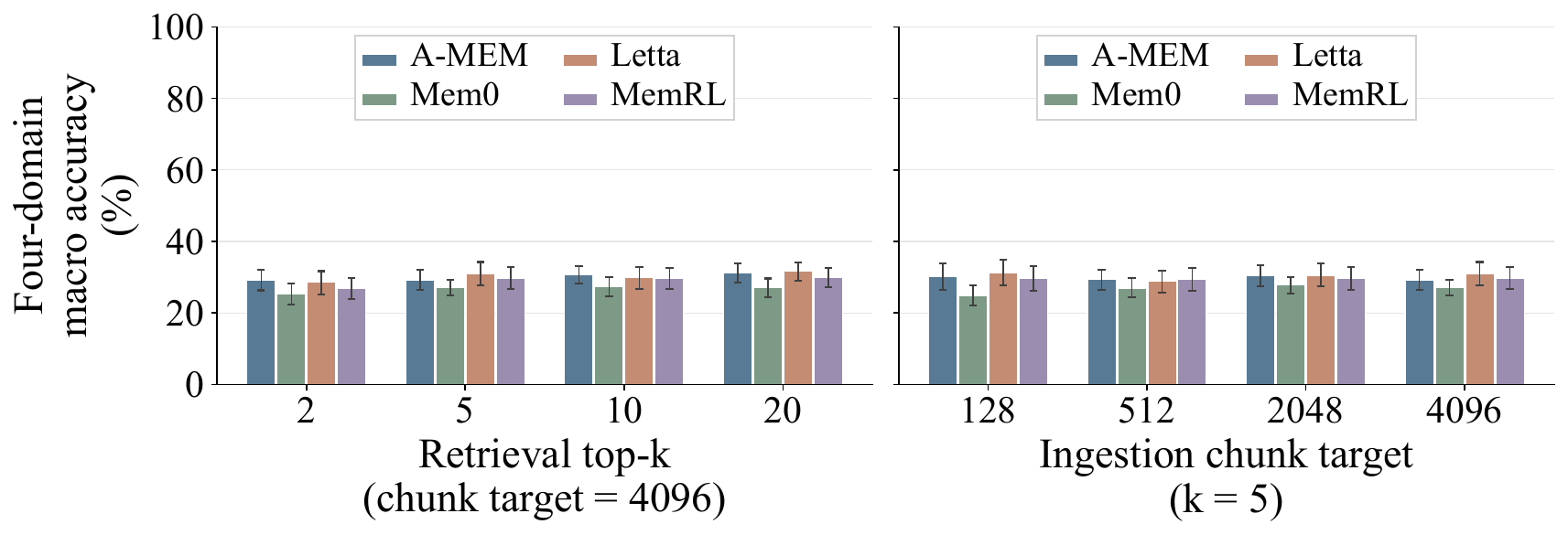}
\par\smallskip
{\small\textbf{(a)} Retrieval and ingestion sensitivity}
\end{minipage}\hfill
\begin{minipage}[t]{0.485\linewidth}
\centering
\includegraphics[width=\linewidth]{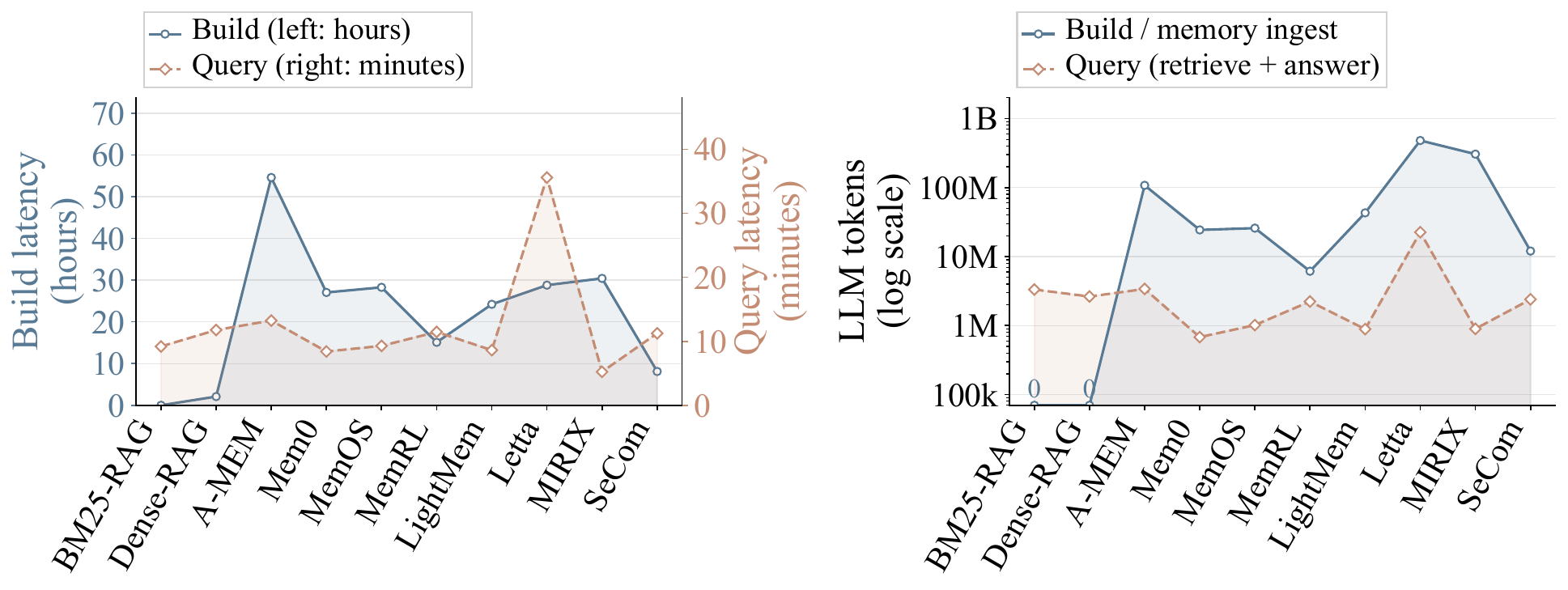}
\par\smallskip
{\small\textbf{(b)} Measured memory costs}
\end{minipage}
\caption{Sensitivity and cost analysis with Qwen3-8B. \textbf{(a)} Four-domain macro accuracy under retrieval top-$k$ and ingestion-chunk ablations, with 95\% within-domain user-bootstrap intervals. \textbf{(b)} Cumulative build/query latency and LLM token usage on the same 32-user cohort. Build and query times are reported in hours and minutes, respectively. Token counts use a logarithmic scale. }
\label{fig:ablation-costs}
\end{figure}

\textbf{Retrieval quantity and ingestion granularity.}
We conduct a separate Qwen3-8B sensitivity study on eight fixed users per domain, covering 32 users and 954 probes, with adapted A-MEM, Mem0, Letta, and MemRL configurations.
We vary retrieval top-$k$ over 2, 5, 10, and 20 at a 4096-token ingestion target, and vary the ingestion target over 128, 512, 2048, and 4096 tokens at $k=5$, using each method's native tokenizer.
Figure~\ref{fig:ablation-costs}(a) reports the equal-domain mean of probe-level accuracies from the recorded run aggregates.
Accuracy ranges from 24.83\% to 31.71\% across the tested settings.
Increasing $k$ from 2 to 20 raises A-MEM from 29.23\% to 31.21\% and Letta from 28.53\% to 31.71\%, while larger ingestion chunks do not consistently improve accuracy.
Thus, varying retrieval quantity or ingestion granularity alone does not remove the observed limitations on this cohort.

\textbf{Memory construction and query costs.}
Figure~\ref{fig:ablation-costs}(b) separates memory construction from subsequent retrieval and answering across ten methods on the same cohort.
The recorded per-user measurements sum to 54.59 hours and 107.80 million LLM tokens for A-MEM construction, compared with 13.24 minutes and 3.39 million tokens for its queries.
This illustrates the substantial upfront cost that memory construction can incur even when subsequent queries are relatively inexpensive.
Time is accumulated per-user latency, not parallel job wall time, and tokens include LLM prompts and completions but exclude embeddings.
Initialization and discarded failed or interrupted runs are excluded.
These measurements characterize the recorded implementations rather than a matched accuracy--cost ranking or an extrapolation to the full dataset.
Separating construction from querying is important because memory is reused across subsequent requests.
A low query cost can conceal substantial processing of the preceding history, while the value of that investment depends on how often the resulting memory is consulted and updated.
The sensitivity and cost analyses therefore address complementary questions: how retrieval and ingestion choices affect decisions, and where computation is spent over the memory lifecycle.

%% file: tables/dual_model_results.tex
\begin{table}[tb]
\centering
\footnotesize
\setlength{\tabcolsep}{1.1pt}
\caption{Main results with Qwen3-235B-A22B and GLM-5.3-Flash: exact-choice accuracy (\%). Macro is the unweighted four-domain mean. Controls measure performance without memory or with compressed history. Retrieval baselines select sessions, while memory methods construct and maintain memory representations. Letta and MIRIX are shown separately as multi-store memory methods. Gray averages weight methods equally. All averages excludes the No Memory and Full Memory controls. TH-RAG denotes Temporal Hybrid RAG.}
\label{tab:main-results}
\renewcommand{\arraystretch}{1.16}
\setlength{\aboverulesep}{0pt}
\setlength{\belowrulesep}{0pt}
\begin{tabularx}{\linewidth}{@{}l|*{5}{>{\centering\arraybackslash}X}|*{5}{>{\centering\arraybackslash}X}@{}}
\toprule
\multirow{2}{*}{Method} & \multicolumn{5}{c|}{\strut\textbf{Qwen3-235B-A22B}} & \multicolumn{5}{c}{\strut\textbf{GLM-5.3-Flash}} \\
 & Food & Finance & Software & Travel & Macro & Food & Finance & Software & Travel & Macro \\
\midrule
\rowcolor{habitGroupBlue}\multicolumn{11}{c}{\strut\itshape Controls} \\
No Memory & 22.38 & 23.54 & 24.85 & 28.31 & 24.77 & 23.81 & 22.95 & 22.50 & 32.46 & 25.43 \\
Full Memory & 48.25 & 20.54 & 21.76 & 33.38 & 30.99 & 45.56 & 33.77 & 37.50 & 51.08 & 41.98 \\
\midrule
\rowcolor{habitGroupBlue}\multicolumn{11}{c}{\strut\itshape Retrieval baselines} \\
Recency-5 & 24.13 & 22.51 & 24.26 & 31.23 & 25.53 & 18.10 & 23.46 & 23.53 & 39.69 & 26.20 \\
Recency-10 & 23.02 & 21.71 & 23.68 & 31.38 & 24.95 & 18.25 & 25.88 & 22.94 & 41.85 & 27.23 \\
BM25 & 45.08 & 23.03 & 22.50 & 39.38 & 32.50 & 31.90 & 27.19 & 24.12 & 48.62 & 32.96 \\
Dense & 39.84 & 22.08 & 23.53 & 39.85 & 31.32 & 30.32 & 24.85 & 24.12 & 53.54 & 33.21 \\
TH-RAG & 45.24 & 22.37 & 22.79 & 38.00 & 32.10 & 28.89 & 26.90 & 24.12 & 49.54 & 32.36 \\
\rowcolor{habitAverageGray}\textbf{Retr. avg.} & \textbf{35.46} & \textbf{22.34} & \textbf{23.35} & \textbf{35.97} & \textbf{29.28} & \textbf{25.49} & \textbf{25.66} & \textbf{23.77} & \textbf{46.65} & \textbf{30.39} \\
\midrule
\rowcolor{habitGroupBlue}\multicolumn{11}{c}{\strut\itshape Memory methods} \\
A-MEM & 42.70 & 22.44 & 25.74 & 41.38 & 33.06 & 27.94 & 28.29 & 25.15 & 53.08 & 33.61 \\
Mem0 & 32.70 & 22.95 & 25.29 & 33.08 & 28.51 & 34.76 & 25.51 & 25.44 & 46.62 & 33.08 \\
MemOS & 40.32 & 21.86 & 21.47 & 34.77 & 29.60 & 36.67 & 24.05 & 24.56 & 47.23 & 33.13 \\
MemRL & 42.70 & 23.32 & 23.24 & 38.15 & 31.85 & 27.46 & 24.93 & 23.82 & 49.54 & 31.44 \\
LightMem & 28.25 & 22.81 & 25.88 & 31.08 & 27.01 & 33.33 & 23.68 & 21.18 & 44.31 & 30.63 \\
SeCom & 41.11 & 22.08 & 22.94 & 38.31 & 31.11 & 28.41 & 24.71 & 22.50 & 52.00 & 31.91 \\
\rowcolor{habitAverageGray}\textbf{Mem. avg.} & \textbf{37.96} & \textbf{22.58} & \textbf{24.09} & \textbf{36.13} & \textbf{30.19} & \textbf{31.43} & \textbf{25.20} & \textbf{23.78} & \textbf{48.80} & \textbf{32.30} \\
\midrule
\rowcolor{habitGroupBlue}\multicolumn{11}{c}{\strut\itshape Multi-store memory} \\
Letta & 41.59 & 22.73 & 23.24 & 40.92 & 32.12 & 27.14 & 25.73 & 24.12 & 54.62 & 32.90 \\
MIRIX & 26.98 & 22.73 & 25.59 & 34.62 & 27.48 & 27.14 & 24.93 & 21.76 & 47.85 & 30.42 \\
\rowcolor{habitAverageGray}\textbf{Multi. avg.} & \textbf{34.29} & \textbf{22.73} & \textbf{24.42} & \textbf{37.77} & \textbf{29.80} & \textbf{27.14} & \textbf{25.33} & \textbf{22.94} & \textbf{51.24} & \textbf{31.66} \\
\midrule
\rowcolor{gray!20}\textbf{All avg.} & \textbf{36.44} & \textbf{22.51} & \textbf{23.86} & \textbf{36.32} & \textbf{29.78} & \textbf{28.49} & \textbf{25.39} & \textbf{23.64} & \textbf{48.35} & \textbf{31.47} \\
\bottomrule
\end{tabularx}
\vspace{-10pt}
\end{table}

%% file: tables/diagnostic_results.tex
\begin{table}[tb]
\centering
\small
\caption{No Memory and private-information diagnostics: probe-level exact-choice accuracy (\%) on the same four datasets. Oracles are non-deployable. Reader protocols differ across models, so this is not a scale-only ablation.}
\label{tab:diagnostic-results}
\begin{tabularx}{\linewidth}{@{}l|*{4}{>{\centering\arraybackslash}X}|*{4}{>{\centering\arraybackslash}X}@{}}
\toprule
\multirow{2}{*}{Method} & \multicolumn{4}{c|}{Qwen3-8B} & \multicolumn{4}{c}{Qwen3-235B-A22B} \\
 & Food & Finance & Software & Travel & Food & Finance & Software & Travel \\
\midrule
No Memory & 20.79 & 23.32 & 27.21 & 26.46 & 22.38 & 23.54 & 24.85 & 28.31 \\
Oracle Evidence & 53.33 & 29.82 & 31.91 & 42.15 & 58.25 & 42.91 & 39.26 & 55.54 \\
Oracle Habit State & 92.54 & 93.71 & 94.56 & 98.31 & 98.41 & 98.54 & 99.26 & 99.23 \\
\bottomrule
\end{tabularx}
\end{table}

%% file: sec/5_conclusion.tex
\section{Conclusion}

HABIT-Bench evaluates whether agents can infer behavioral defaults from longitudinal weak evidence and use them to make context-appropriate decisions. 
Its four domains require agents to integrate distributed observations while tracking applicability, exceptions, temporal validity, and user endorsement. Across the evaluated configurations, memory benefits vary substantially by domain, and richer memory pipelines do not consistently outperform retrieval or budget-feasible history compression. Oracle controls show that supplying annotated evidence still leaves substantial decision difficulty, while evidence-attribution diagnostics distinguish partial retrieval from grounded action selection. These findings motivate memory representations that preserve the relationships between a default, its supporting observations, and its conditions of validity. HABIT-Bench provides a controlled setting for studying how such representations support context-appropriate decisions.

%% file: sec/6_appendix.tex
\section{Domain-specific Construction and Validation}
\label{app:domain-construction}

The four domains share chronological user histories and a deterministic four-choice interface, while differing in their source tasks, evidence structures, and habit composition.
The following sections describe the construction and validation of the evaluated datasets.

\subsection{Food}

\paragraph{Templates and source compatibility.}
The Food pipeline starts with 20 habit templates for private preparation, finishing, and serving choices. Each template specifies its applicability condition, default action, a competing action, boundary condition/action, and exception condition/action. Illustrative axes include the placement of herbs, chili, and briny accents, as well as recipe finishing styles. Templates that are effectively generic safety rules are filtered out. Trigger patterns constrain assignment to compatible recipe seeds, so that an oven-related habit is not inserted into an unrelated cooking task. The dialogue and recipe seeds come from CookDial~\citep{jiang2023cookdial}. User continuity, habit assignments, and probe labels are constructed within HABIT-Bench rather than inherited from CookDial.

\paragraph{Evidence scheduling.}
Each user receives seven habits. A stable habit has ten signal sessions: one candidate shortlist, one distant ordinal resolution, three ratified reuses, two boundary observations, one one-case exception, one nonresolving rehearsal, and one rejected assistant suggestion. Key signals are distributed across early, middle, and late history. The boundary observations deliberately include a weak positive and an adjacent near-boundary negative: the contrast is locally available even though the overall habit evidence is longitudinal. This is not a claim that every supporting observation is far from every other observation. Sixty ordinary sessions are added to the 70 signal sessions, giving 130 sessions per user and 3,900 sessions in total.

\paragraph{Naturalization and probes.}
Naturalization rewrites planned sessions into cooking conversations while preserving recipe facts and habit evidence.
Validation checks turn limits, protected recipe numbers, evidence-index alignment, preference leakage in ordinary sessions, mechanical phrasing, and benchmark-related terms.
Each of the 210 user-level habits yields direct-use, boundary, and exception probes.
Query surfaces reduce direct overlap with reuse evidence, and all four alternatives are adapted to the current scenario so that the gold action is not the only physically applicable option.
The final dataset contains 630 probes.

\paragraph{Exception scope and validation.}
Exceptions authorize a local departure from the default rather than a durable replacement.
The decision target distinguishes applying the default, respecting its boundary, and following the exception in its applicable context.
Naturalization and human validation are part of the construction pipeline, with review criteria and results reported in Appendix~\ref{app:human-audit}.

\subsection{Finance}

\paragraph{Source transformation and persona consistency.}
The Finance source from MultiDoGO~\citep{peskov2019multi} contains 8,832 conversations and 157,229 utterances.
Conversations are reconstructed by identifier and turn order, then assigned a primary task intent.
They supply task types, slot categories, interaction patterns, and style exemplars, while original identities and sensitive financial fields are replaced in the synthetic lifelines.
The resulting 17,405 source-linked sessions use 7,049 primary source conversations, with no primary seed used more than three times.
Source-linked and evidence-annotated sessions are overlapping categories.

Each of the 36 users has a stable banking persona, including synthetic account identifiers and a consistent financial background. The history has 540 sessions per user, spanning January 2021 through June 2025. The final 19,440 sessions include 16,180 without memory annotations and 3,260 with some annotation; annotation is not synonymous with decisive evidence.

\paragraph{Habit state and evidence topology.}
There are eight candidate habits. Each user activates six and leaves two absent; two active habits undergo replacement. Workflow variants cover money movement, card/account changes, minimal secure verification, urgent fraud handling, account-review presentation, dispute preparation, noncommittal credit/loan analysis, and current payment-status checks. The secure-verification habit is active for every Finance user. Variants are designed to distinguish user-specific processes rather than safe versus plainly unsafe alternatives.

A standing policy is recovered by joining an ordered shortlist with a distant user resolution that refers to an ordinal alternative. For drift habits, an additional shortlist/resolution pair establishes a replacement. One-off cases, unaccepted assistant proposals, unresolved rehearsals, and tentative signals are not automatically binding. Probe evidence must link each task component to the appropriate workstream, policy pair, user resolution, and target time. The private chains distinguish decisive, temporal-context, and nonbinding evidence.

\paragraph{Probe construction and validation.}
The 1,368 probes comprise 944 two-habit and 424 three-habit decisions.
Types include as-of reversals, interleaved temporal composition, surface-decoy pairs, reference-case reconstruction, suggestion rejection, provenance-sensitive composition, and scope/temporal pairs.
Validation checks gold--graph consistency, query--choice completeness, evidence topology, and surface quality.
In particular, choices must cover every required task component, with no component supplied only by the gold alternative.
The dataset also defines 288 decision units for complementary unit-level aggregation.
Human review retains 336 of 340 sampled items and identifies four query--choice coverage defects, which are corrected before final evaluation (Appendix~\ref{app:human-audit}).

\subsection{Software}

\paragraph{Source transformation and environments.}
The Software source from MultiDoGO~\citep{peskov2019multi} contains 5,562 conversations and 86,220 utterances.
Task intents are inferred from complete conversations.
The resulting 8,678 source-linked sessions use 2,828 primary conversations, with at most four realizations per primary seed.
They retain support-task structure, slot categories, clarification patterns, and expression styles while replacing original company identities, credentials, contact fields, and repetitive helpdesk language.
Each synthetic user has a consistent organization, operating system, browser, product suite, support account, and work context.

\paragraph{Habit assignment and temporal interference.}
Eighteen users each have 540 sessions, giving 9,720 sessions: 8,014 without memory annotations and 1,706 with some annotation. Seventeen users have six active habits and one has five. The seven candidate habits concern diagnostics, documentation lookup, escalation thresholds, secure account recovery, platform-specific instructions, backup before risky changes, and receipts after actual ticket submission.

Software shares Finance's distant shortlist/resolution mechanism and contains 71 replacement chains.
Drift frequency varies across users: nine have two replacements each, eight have six, and one has five.
Competing evidence includes similar errors in different environments, case-local diagnostic routes, unaccepted proposals, and prior unresolved checkpoints. Ticket receipts are scoped to submitted tickets, not merely drafts or ongoing troubleshooting.

\paragraph{Probe construction and validation.}
The 680 probes comprise 464 two-habit and 216 three-habit decisions.
They combine workflow selection with temporal state, provenance, and reference-case continuation.
Validation checks query--choice coverage, gold--graph consistency, and nonredundant query text.
The dataset defines 161 decision units, and human review accepts all 311 sampled items (Appendix~\ref{app:human-audit}).

\subsection{Travel}

\paragraph{Source-derived motifs and history-first realization.}
Travel begins with independent flight and hotel conversations from Taskmaster-2~\citep{byrne2019taskmaster}.
Source conversations provide behavioral motifs rather than cross-conversation user identities.
Semantic consolidation yields 47 grounded instances, 18 policy prototypes across 13 habit families, and 194 habits assigned to 30 synthetic users.

Each user is specified by a dossier and chronological arc.
Natural multi-turn sessions are authored before evidence annotation and blind recoverability checks, so the history establishes the habit before the probe tests its application.
The histories contain 4,092 sessions, with 124--147 per user: 790 evidence-mode and 3,302 ordinary-mode sessions.
There are 194 initial habit states and eight revisions within the lifelines.
Probe-visible sessions are restricted by their chronological cutoffs.

\paragraph{Decision probes.}
The dataset contains 650 probes: 456 single-habit and 194 two-habit decisions.
Probe families include cross-context transfer, conflict resolution, evidence disambiguation, exception/default composition, scope composition, and temporal-state composition.
Private annotations specify the relevant habits and evidence for each decision.
The query and alternatives provide current candidate facts, while historical sessions establish the user-specific behavioral rule needed to choose among them.

\paragraph{Validation and evaluation scope.}
Validation combines gold-blind, history-informed answerability checks with history-free checks for generic-best shortcuts.
A and B independently assess the sampled items, C adjudicates disagreements, and content revisions receive targeted blind review.
Appendix~\ref{app:human-audit} reports agreement and review coverage.
The experiments evaluate the full probe set with user-isolated memories and chronological cutoffs.
Travel is a controlled diagnostic set, not an independently sequestered generalization test.

\subsection{Cross-domain Interpretation}

The common four-choice interface does not make the evidence mechanisms identical. Food includes both split decisions and repeated reuse, with local contrast for boundaries. Finance and Software predominantly test cross-session decision reconstruction and policy-state maintenance. Travel emphasizes behavioral-rule recovery and compositional application. The benchmark therefore evaluates integration of incomplete longitudinal evidence, not exclusively statistical inference from unratified behavior. Likewise, deterministic action labels do not establish a probabilistic habit-calibration task. These distinctions should constrain both the capability claims and the interpretation of domain-averaged accuracy.

\section{Human Validation Protocol and Results}
\label{app:human-audit}

\subsection{Gold-blind Review and Quality Criteria}

\begin{figure}[htbp]
\centering
\includegraphics[width=\linewidth]{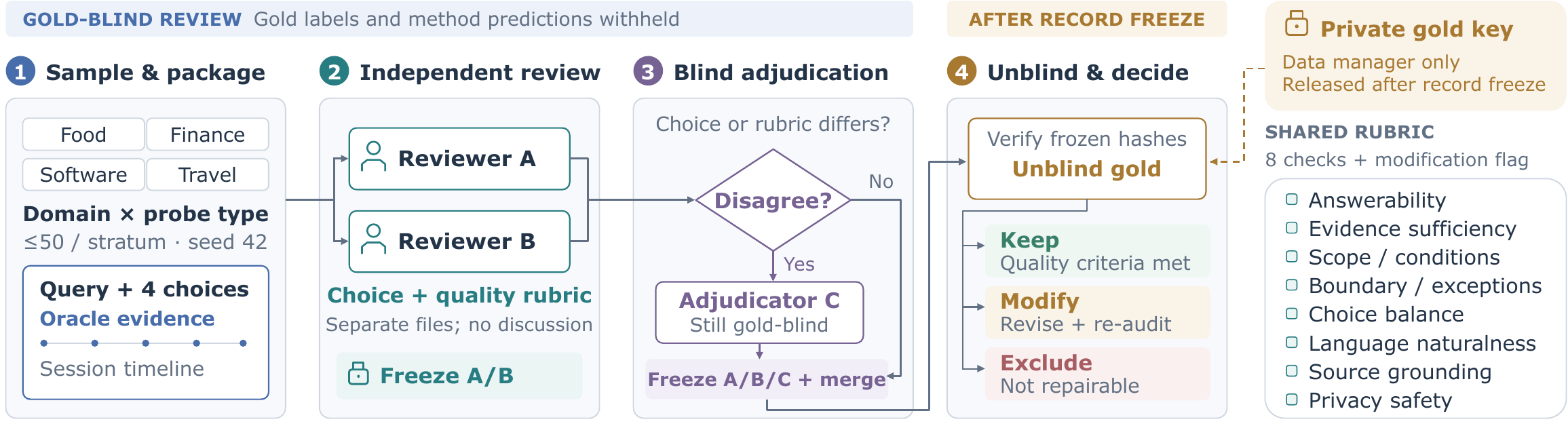}
\caption{Human-validation protocol. A and B independently choose an action and apply the quality rubric. C adjudicates answer or quality disagreements without gold access. Only after the merged record is frozen does the data manager compare it with gold and assign a release disposition. Content corrections receive new, versioned inputs and targeted blind review. The diagram describes the protocol, while the tables report the completed stages and coverage in each domain.}
\label{fig:human-audit-protocol}
\end{figure}

The review separates three questions: whether the evidence supports a unique action, whether the item meets the construction rubric, and whether its gold label agrees with the independently recovered action.
Figure~\ref{fig:human-audit-protocol} summarizes the process.
Reviewers receive anonymous queries, ordered alternatives, and chronological oracle evidence packets containing decisive evidence, necessary temporal context, and relevant nonbinding observations.
Private habit graphs, gold choices, method predictions, and the other reviewer's initial judgments are withheld during independent annotation.
The evidence packets are selected using the private annotations, not retrieved by a tested memory method.
This design tests decision recoverability conditional on those packets and does not measure human retrieval from the full lifelines.

A and B record a selected action, eight quality judgments, and a separate modification flag with a reason.
An unanswerable or ambiguous item is left without a selected choice.
Inapplicable rubric fields are excluded from their field-specific denominators rather than counted as passes.
After A/B records are frozen, any difference in the choice or a quality field is routed to C.
Agreed judgments are retained, and C resolves the disputed judgments before gold is revealed for scoring.
Thus, the three-person protocol does not mean that each item receives three independent initial annotations.

The release decision is \emph{Keep} when the item passes all applicable criteria without a required modification, \emph{Modify} for a locally repairable defect, and \emph{Exclude} when the item is invalid or requires reconstruction.
A gold mismatch is not resolved by forcing the reviewer to match the stored label.
Corrections create a new data revision while preserving the original judgments, and content changes require targeted blind re-review.
Table~\ref{tab:human-audit-rubric} summarizes quality judgments and validation outcomes across all four domains.
The rubric requires a unique evidence-supported answer, correct contextual and temporal scope, and appropriate treatment of boundaries, exceptions, and nonbinding evidence.
It also checks that alternatives have comparable task coverage without generic-best or surface shortcuts, that language is coherent, that decisions are grounded in valid evidence rather than unratified suggestions, and that no real sensitive identifiers are exposed.
Evidence sufficiency alone is not sufficient for acceptance if no alternative fulfills the complete request.

\input{tables/human_audit_rubric.tex}

\subsection{Sampling and Agreement}

Sampling is stratified by domain and probe type, using seed 42 and at most 50 items per stratum, with smaller strata included in full.
Additional targeted items are included when required by a repair manifest.
We report the reviewed sample rather than extrapolating its pass rate to the domain population, because strata have different sampling fractions.
Answer agreement and Cohen's $\kappa$ use pre-adjudication responses on items with valid choices from both A and B.
Gold agreement uses each reviewer's valid choices as its denominator, with blank responses reported separately.
Quality-field agreement is distinct from answer agreement, and a constant binary field can have undefined $\kappa$ despite complete raw agreement.
Table~\ref{tab:human-audit-agreement} reports independent A/B results for all four domains before adjudication or label correction.
Domain-specific agreement scores are reported separately, and adjudicated labels are not substituted into these independent agreement measurements.

\input{tables/human_audit_agreement.tex}

Finance has 340 jointly annotated items across seven probe types.
Five additional planned items lack B annotations and are excluded for coverage, not classified as quality failures.
This leaves three of eight requested targeted items within the double-annotated sample.
The 336 jointly valid A/B choices all agree, but four further items lack an alternative satisfying an additional output-placement requirement.
C marks these four as \emph{Modify}, yielding 336 Keep, four Modify, and no Exclude decisions at the audit stage.
The 336 valid adjudicated choices all match gold.
The evidence for the intended habit is sufficient in all four cases.
The modification flags instead identify an extra output-placement requirement omitted from every alternative, illustrating that the rubric checks the full request rather than only agreement with the intended gold choice.
We retain these judgments in the audit instead of counting the affected items as gold mismatches or forcing reviewers to choose the nearest alternative.
High agreement on the valid choices therefore coexists with a small number of strict quality flags.
The four local issues were subsequently corrected in the dataset used for final evaluation.
Table~\ref{tab:human-audit-rubric} preserves their original audit dispositions to show the sensitivity of the quality checks, rather than to indicate outstanding defects in the evaluated dataset.

Software has 311 jointly annotated items across seven probe types, including all four requested targeted items.
A and B agree on 308 of 309 jointly valid choices.
One answer disagreement and two blank B responses produce three adjudication cases.
C resolves these cases, yielding 311 Keep decisions and 311/311 adjudicated choices matching gold.
The adjudications distinguish historical workflow state from a current request and distinguish neutral packet identifiers from actual output constraints.

\subsection{Food Review}

The Food review covers 200 probes and follows the same answerability and quality rubric.
Naturalization is a completed construction stage that improves conversational fluency while preserving the habit evidence and decision constraints.
Table~\ref{tab:human-audit-rubric} reports 200/200 naturalness passes and 0/200 items requiring modification.
Independent A/B answer agreement is 99.50\%, with 200/200 A--gold and 199/200 B--gold matches (Table~\ref{tab:human-audit-agreement}).

\subsection{Travel Validation}

Travel's independent review covers 618 probes, with answer agreement and blank responses reported in Table~\ref{tab:human-audit-agreement}.
C adjudicates 198 items with at least one decision-field disagreement.
Content revisions undergo targeted gold-blind validation, and all 79 revised items pass this stage.
These targeted checks assess the revised content, rather than provide a second independent A/B agreement estimate.
Table~\ref{tab:human-audit-travel} summarizes coverage within the 650-probe dataset, including the 32 probes outside the human-review sample.
The boundary/exception criterion applies to 495 reviewed items, with 123 inapplicable entries excluded from its denominator.

\input{tables/human_audit_travel.tex}

\section{Implementation Provenance}
\label{app:implementation-provenance}

A-MEM, Letta, LightMem, Mem0, MemOS, MemRL, and MIRIX are evaluated through MedMemoryBench adapters \citep{wang2026medmemorybench}.
These are \emph{official-code benchmark adapters}: they retain the pinned source's memory lifecycle but expose retrieval-only context to the common HABIT answerer, so they are not upstream-native end-to-end runs.
SeCom is an \emph{official-source adapted baseline} configured for chronological ingestion.
The session-retrieval methods are HABIT-native retrieval baselines; No Memory, Full Memory, and Full History are evaluator controls; and Oracle Evidence and Oracle Habit State are private diagnostic oracles rather than deployable systems.

\input{tables/method_provenance.tex}

\Needspace{0.45\textheight}
\section{Capability-Slice Matrix}
\label{app:capability-slices}

\input{tables/capability_results.tex}

\Needspace{6\baselineskip}
\section{Extended Model Evaluation}
\label{app:model-scaling}

\subsection{Qwen3-8B: Memory Methods and Retrieval Baselines}
\label{app:qwen8b}

\input{tables/qwen8b_combined_results.tex}

Table~\ref{tab:qwen8b-combined} brings together the Qwen3-8B memory-method and session-retrieval results under one answer configuration.
It reports user-macro accuracy and available user-cluster bootstrap intervals, rather than the probe-level statistic used in the main experiment.
Full Memory reaches 41.1\% on Food, while BM25 RAG reaches 42.7\%.
The leading memory-method scores are 23.1\% on Finance with Mem0 and SeCom, 27.5\% on Software with MIRIX, and 34.0\% on Travel with A-MEM and Letta.
These results provide supplementary context for the main evaluation without mixing models or aggregation levels. 

\input{tables/diagnostic_user_macro.tex}

Table~\ref{tab:diagnostic-user-macro} preserves the user-macro view of the Qwen3-8B controls and their user-cluster intervals.
It complements the probe-level comparison in Table~\ref{tab:diagnostic-results} without treating the two aggregation levels as interchangeable.

\subsection{Qwen Model Comparison}

\input{tables/qwen_scale_micro.tex}

Table~\ref{tab:qwen-scale-micro} provides the exact values plotted in Figure~\ref{fig:qwen-scaling}, using the four-domain mean of within-domain probe-level accuracy.
This matches the aggregation in the main experiment and differs from averaging the user-macro values in Table~\ref{tab:qwen8b-combined}.
All columns use the same Food, Finance, Software, and Travel datasets.
The comparison changes both the model used by LLM-dependent memory stages and the answer model.
In addition, the Qwen3-235B-A22B results use a revised reader protocol, so the observed differences are descriptive rather than a controlled estimate of the effect of parameter count.

\subsection{Model Configurations and Decision Protocol}
\label{app:decision-protocol}

\input{sec/appendix_model_protocol}

\subsection{GLM-5.3-Flash: Results and Protocol}
\label{app:glm53flash}

\begin{figure}[htbp]
\centering
\includegraphics[width=\linewidth]{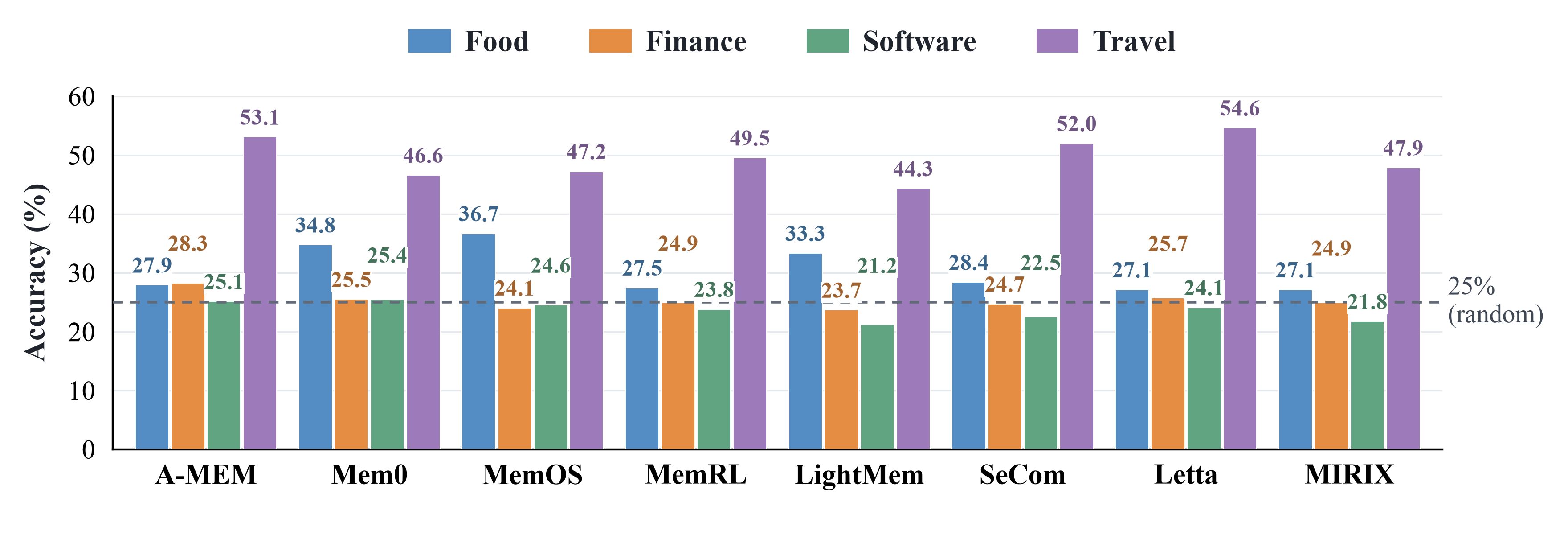}
\vspace{-30pt}
\caption{Memory-method accuracy with GLM-5.3-Flash across four domains. Bars show probe-level accuracy, and the dashed line marks 25\% uniform random choice. Full results, including controls and retrieval baselines, appear in Table~\ref{tab:main-results}.}
\label{fig:glm-domain}
\end{figure}

Figure~\ref{fig:glm-domain} visualizes the eight memory methods from the GLM-5.3-Flash block of Table~\ref{tab:main-results}.

The model and inference settings are given in Appendix~\ref{app:decision-protocol}.
Pooled probe-level Wilson intervals, where reported, are distinct from user-cluster intervals and repeated-run variability and are not used for domain-level significance claims.

\subsection{Qwen3-32B: Evidence Retrieval and Grounded Answers}
\label{app:retrieval-diagnostics}

Table~\ref{tab:qwen32b-retrieval} reports detailed metrics with Qwen3-32B for methods that expose source-session attribution.
LightMem and MIRIX are excluded from this evidence-specific comparison because their memory outputs do not provide stable source-session lineage, not because their retrieved text is necessarily irrelevant.
Their answer accuracies remain in the model-size comparison.
Full Memory is a separate context-coverage control, rather than a top-five retrieval method.

\begingroup
\setlength{\intextsep}{5pt plus 2pt minus 2pt}
\input{tables/qwen32b_retrieval_diagnostics.tex}
\endgroup

For a probe, let $E_p$ be its annotated positive evidence sessions and $S_p^5$ the first five unique source-session identifiers attributed by the method.
\textbf{ER@5} is the fraction of $E_p$ covered by $S_p^5$, averaged over eligible probes.
\textbf{Hit@5} measures whether at least one positive evidence session is retrieved.
\textbf{NDCG@5} measures binary-relevance ranking quality with logarithmic rank discounting, and \textbf{MRR@5} is the reciprocal rank of the first positive hit, or zero when there is none.
\textbf{Joint A+E} requires both a correct answer and an evidence hit.
\textbf{Clean Grounded@5} additionally requires that none of the first five attributed sessions is annotated as nonbinding evidence.
Each metric uses its annotation-eligible probes, so unavailable fields are not converted to zero.
In particular, Clean Grounded@5 uses probes with nonbinding-evidence annotations and does not include Food.
For Full Memory, ER measures evidence coverage in the supplied context and Joint A+E uses a hit anywhere in that context, without a five-session cutoff.

For a nonempty annotated evidence set $E_p$, the per-probe bound is $\mathrm{ER@5}_p\leq\min(5,|E_p|)/|E_p|$. Coverage therefore depends on the size of the annotated set, which is not itself a specification of a minimal sufficient reasoning chain. Joint A+E requires at least one evidence hit rather than complete support for every decision component. Clean Grounded@5 additionally imposes the absence of annotated nonbinding sessions from the attributed context; retrieving such a session can still be useful for determining that a proposal was never accepted. We retain these as complementary context-and-answer diagnostics rather than direct measures of improper personalization.

The ranked methods have ER@5 between 11.17\% and 13.89\%, despite Hit@5 between 37.77\% and 41.44\%.
Retrieving some relevant evidence therefore does not imply reconstructing the full support for a habit.
Joint A+E ranges from 12.92\% to 16.20\%, while Clean Grounded@5 ranges from 5.49\% to 8.20\% on its eligible subset.
These measures diagnose different aspects of evidence access and use, and their different denominators preclude interpreting the difference as a causal error decomposition.

\input{sec/appendix_case_studies}

%% file: tables/human_audit_rubric.tex
\begin{table}[tbp]
\centering
\caption{Human-validation quality judgments across four domains. Counts give positive judgments over applicable reviewed items. Lower is better for the modification flag. Finance retains four initial Modify judgments for output-placement omissions corrected before final evaluation. Completed naturalization and content validation are included in the quality outcomes.}
\label{tab:human-audit-rubric}
\small
\setlength{\tabcolsep}{4pt}
\renewcommand{\arraystretch}{1.18}
\begin{tabularx}{\linewidth}{@{}Xrrrr@{}}
\toprule
Criterion & Food & Finance & Software & Travel \\
\midrule
\rowcolor{habitGroupBlue}\multicolumn{5}{c}{\textit{Recorded quality judgments}} \\
Unique answerability & 200/200 & 336/340 & 311/311 & 618/618 \\
Evidence sufficiency & 200/200 & 340/340 & 311/311 & 618/618 \\
Scope correctness & 200/200 & 336/340 & 311/311 & 618/618 \\
Boundary / exception & 200/200 & 340/340 & 311/311 & 495/495 \\
Choice balance & 200/200 & 340/340 & 311/311 & 618/618 \\
Naturalness & 200/200 & 340/340 & 311/311 & 618/618 \\
Source grounding & 200/200 & 340/340 & 311/311 & 618/618 \\
Privacy safety & 200/200 & 340/340 & 311/311 & 618/618 \\
\midrule
Needs modification & 0/200 & 4/340 & 0/311 & 0/618 \\
\midrule
\rowcolor{habitGroupBlue}\multicolumn{5}{c}{\textit{Review and revision outcomes}} \\
Outcome & \shortstack[r]{Naturalization\\completed} & \shortstack[r]{336 Keep\\4 Modify (corrected)} & 311 Keep & \shortstack[r]{79 repaired\\items accepted} \\
\bottomrule
\end{tabularx}
\end{table}

%% file: tables/human_audit_agreement.tex
\begin{table}[tbp]
\centering
\caption{Independent A/B answer agreement across four domains. $N$: jointly annotated items. $n$: jointly valid choices. A--gold and B--gold give matches over valid choices. Agreement and Cohen's $\kappa$ are measured before adjudication, and gold agreement is evaluated before label corrections.}
\label{tab:human-audit-agreement}
\small
\setlength{\tabcolsep}{5pt}
\begin{tabular}{lrrrrrrr}
\toprule
Domain & $N$ & $n$ & Agreement (\%) & $\kappa$ & A--gold & B--gold & Blanks A/B \\
\midrule
Food & 200 & 200 & 99.50 & 0.9933 & 200/200 & 199/200 & 0/0 \\
Finance & 340 & 336 & 100.00 & 1.0000 & 340/340 & 336/336 & 0/4 \\
Software & 311 & 309 & 99.68 & 0.9957 & 310/311 & 309/309 & 0/2 \\
Travel & 618 & 545 & 92.84 & 0.9044 & 515/577 & 501/564 & 41/54 \\
\bottomrule
\end{tabular}
\end{table}

%% file: tables/human_audit_travel.tex
\begin{table}[tbp]
\centering
\caption{Human-review coverage and targeted content validation in Travel. The coverage rows partition the dataset. The content-validation set is a subset of the reviewed probes.}
\label{tab:human-audit-travel}
\small
\begin{tabular}{@{}lr@{}}
\toprule
Coverage & Probes \\
\midrule
Distinct reviewed probes & 618 \\
Outside the human-review sample & 32 \\
Total probes & 650 \\
\midrule
Content-revised probes passing targeted validation & 79/79 \\
\bottomrule
\end{tabular}
\end{table}

%% file: tables/method_provenance.tex
\begin{table}[ht]
\caption{Implementation provenance of evaluated systems. Adapters share HABIT-Bench chronology, retrieval, and answer interfaces; they are not upstream-native end-to-end evaluations.}
\label{tab:implementation-provenance}
\centering
\small
\begin{tabularx}{\linewidth}{p{0.30\linewidth}X}
\toprule
Implementation class & Methods \\
\midrule
official-code adapter & A-MEM, Letta, LightMem, Mem0, MemOS, MemRL, MIRIX \\
official-source adapted & SeCom \\
HABIT-native retrieval & Recency-5, Recency-10, BM25 RAG, Dense RAG, Temporal Hybrid RAG \\
evaluator control & Full History, Full Memory, No Memory \\
private diagnostic oracle & Oracle Evidence, Oracle Habit State \\
\bottomrule
\end{tabularx}
\end{table}

%% file: tables/capability_results.tex
\begin{table}[ht]
\caption{Qwen3-8B capability slices: probe-level exact-choice accuracy (\%) in Finance, Software, and Travel. Finance/Software columns pool the two domains. The single-probe Travel boundary slice is omitted from the columns (no method answers it correctly).}
\label{tab:capabilities}
\centering
\scriptsize
\setlength{\tabcolsep}{3.2pt}
\begin{tabular}{lrrrrr}
\toprule
Method & F/S temporal & F/S provenance & F/S reference & Travel induction & Travel exception \\
\midrule
Full Memory & 22.8 & 20.2 & 21.2 & 29.6 & 39.4 \\
A-MEM & 24.8 & 23.0 & 23.1 & 35.2 & 36.4 \\
Letta & 23.5 & 22.0 & 23.1 & 37.0 & 45.5 \\
LightMem & 24.4 & 22.8 & 22.5 & 37.0 & 27.3 \\
Mem0 & 25.1 & 24.0 & 22.8 & 35.2 & 42.4 \\
MemOS & 24.6 & 22.8 & 22.2 & 35.2 & 27.3 \\
MemRL & 23.3 & 21.7 & 24.4 & 38.9 & 36.4 \\
MIRIX & 23.5 & 24.9 & 24.4 & 37.0 & 27.3 \\
SeCom & 23.4 & 23.4 & 24.7 & 38.9 & 39.4 \\
\bottomrule
\end{tabular}
\end{table}

%% file: tables/qwen8b_combined_results.tex
\begin{table}[t]
\caption{Supplementary Qwen3-8B comparison. Values are user-macro accuracy (\%) with 95\% user-cluster intervals in gray. Full Memory is a control. Shaded averages weight methods equally within each group or across all methods and exclude controls. Means are computed before display rounding. Intervals are not averaged.}
\label{tab:qwen8b-combined}
\centering
\scriptsize
\setlength{\tabcolsep}{3.5pt}
\renewcommand{\arraystretch}{1.10}
\begin{tabularx}{\linewidth}{l*{4}{>{\centering\arraybackslash}X}}
\toprule
Method & Food & Finance & Software & Travel \\
\midrule
\rowcolor{habitGroupBlue}\multicolumn{5}{c}{\strut\itshape Controls} \\
Full Memory & 41.1 \textcolor{habitCI}{[37.6, 44.8]} & 20.8 \textcolor{habitCI}{[18.1, 23.6]} & 23.2 \textcolor{habitCI}{[20.1, 26.1]} & 28.1 \textcolor{habitCI}{[24.0, 32.7]} \\
\midrule
\rowcolor{habitGroupBlue}\multicolumn{5}{c}{\strut\itshape Retrieval baselines} \\
Recency-5 & 21.3 \textcolor{habitCI}{[17.8, 25.1]} & 23.0 \textcolor{habitCI}{[20.5, 25.7]} & 25.9 \textcolor{habitCI}{[23.1, 28.7]} & 26.5 \textcolor{habitCI}{[22.1, 31.0]} \\
Recency-10 & 21.9 \textcolor{habitCI}{[19.2, 24.8]} & 22.8 \textcolor{habitCI}{[20.2, 25.5]} & 26.3 \textcolor{habitCI}{[23.5, 29.3]} & 30.2 \textcolor{habitCI}{[25.5, 34.9]} \\
BM25 RAG & 42.7 \textcolor{habitCI}{[39.5, 45.7]} & 23.0 \textcolor{habitCI}{[20.7, 25.4]} & 24.4 \textcolor{habitCI}{[21.2, 27.6]} & 33.6 \textcolor{habitCI}{[29.0, 38.2]} \\
Dense RAG & 39.8 \textcolor{habitCI}{[37.0, 42.7]} & 22.8 \textcolor{habitCI}{[20.3, 25.4]} & 25.4 \textcolor{habitCI}{[22.4, 28.6]} & 33.5 \textcolor{habitCI}{[29.8, 37.1]} \\
Temporal Hybrid RAG & 41.7 \textcolor{habitCI}{[37.9, 45.6]} & 21.1 \textcolor{habitCI}{[18.5, 23.7]} & 26.3 \textcolor{habitCI}{[22.6, 30.2]} & 33.5 \textcolor{habitCI}{[29.2, 37.8]} \\
\rowcolor{habitAverageGray}\textbf{Retr. avg.} & \textbf{33.5} & \textbf{22.5} & \textbf{25.7} & \textbf{31.5} \\
\midrule
\rowcolor{habitGroupBlue}\multicolumn{5}{c}{\strut\itshape Memory methods} \\
A-MEM & 39.5 \textcolor{habitCI}{[36.0, 43.2]} & 22.6 \textcolor{habitCI}{[20.1, 25.2]} & 26.5 \textcolor{habitCI}{[23.8, 29.3]} & 34.0 \textcolor{habitCI}{[30.2, 37.8]} \\
Mem0 & 29.8 \textcolor{habitCI}{[27.0, 32.7]} & 23.1 \textcolor{habitCI}{[20.4, 26.1]} & 26.8 \textcolor{habitCI}{[23.4, 29.9]} & 27.5 \textcolor{habitCI}{[24.5, 30.8]} \\
MemOS & 38.1 \textcolor{habitCI}{[34.8, 41.6]} & 22.2 \textcolor{habitCI}{[19.3, 25.1]} & 26.3 \textcolor{habitCI}{[23.2, 29.4]} & 31.7 \textcolor{habitCI}{[27.5, 35.9]} \\
MemRL & 37.6 \textcolor{habitCI}{[34.1, 41.1]} & 21.9 \textcolor{habitCI}{[19.2, 24.7]} & 25.0 \textcolor{habitCI}{[21.9, 28.0]} & 33.0 \textcolor{habitCI}{[29.9, 35.9]} \\
LightMem & 32.7 \textcolor{habitCI}{[29.2, 36.2]} & 22.5 \textcolor{habitCI}{[19.8, 25.3]} & 25.5 \textcolor{habitCI}{[21.9, 29.0]} & 28.6 \textcolor{habitCI}{[25.1, 32.4]} \\
SeCom & 38.6 \textcolor{habitCI}{[35.6, 41.6]} & 23.1 \textcolor{habitCI}{[20.5, 25.7]} & 24.9 \textcolor{habitCI}{[21.8, 28.1]} & 33.4 \textcolor{habitCI}{[29.9, 37.0]} \\
\rowcolor{habitAverageGray}\textbf{Mem. avg.} & \textbf{36.1} & \textbf{22.6} & \textbf{25.8} & \textbf{31.4} \\
\midrule
\rowcolor{habitGroupBlue}\multicolumn{5}{c}{\strut\itshape Multi-store memory} \\
Letta & 38.6 \textcolor{habitCI}{[35.1, 41.9]} & 21.8 \textcolor{habitCI}{[19.0, 24.5]} & 25.3 \textcolor{habitCI}{[22.4, 28.1]} & 34.0 \textcolor{habitCI}{[29.7, 38.1]} \\
MIRIX & 28.7 \textcolor{habitCI}{[25.9, 31.4]} & 22.6 \textcolor{habitCI}{[20.0, 25.2]} & 27.5 \textcolor{habitCI}{[24.8, 30.0]} & 28.2 \textcolor{habitCI}{[25.0, 31.4]} \\
\rowcolor{habitAverageGray}\textbf{Multi. avg.} & \textbf{33.7} & \textbf{22.2} & \textbf{26.4} & \textbf{31.1} \\
\midrule
\rowcolor{habitAverageGray}\textbf{All avg.} & \textbf{34.7} & \textbf{22.5} & \textbf{25.9} & \textbf{31.4} \\
\bottomrule
\end{tabularx}
\end{table}

%% file: tables/diagnostic_user_macro.tex
\begin{table}[t]
\centering
\scriptsize
\setlength{\tabcolsep}{3.5pt}
\caption{Qwen3-8B controls and private-information diagnostics under user-macro aggregation, with 95\% user-cluster bootstrap intervals in gray. These intervals describe user resampling, not repeated-run variability.}
\label{tab:diagnostic-user-macro}
\begin{tabular}{lcccc}
\toprule
Method & Food & Finance & Software & Travel \\
\midrule
No Memory & 20.8 \textcolor{habitCI}{[17.1, 24.6]} & 23.3 \textcolor{habitCI}{[20.5, 26.3]} & 27.2 \textcolor{habitCI}{[23.9, 30.3]} & 26.9 \textcolor{habitCI}{[23.4, 30.4]} \\
Oracle Evidence & 53.3 \textcolor{habitCI}{[50.3, 56.5]} & 29.8 \textcolor{habitCI}{[27.4, 32.3]} & 31.9 \textcolor{habitCI}{[29.1, 34.7]} & 42.2 \textcolor{habitCI}{[37.8, 46.4]} \\
Oracle Habit State & 92.5 \textcolor{habitCI}{[90.6, 94.4]} & 93.7 \textcolor{habitCI}{[92.4, 94.9]} & 94.6 \textcolor{habitCI}{[92.2, 96.6]} & 98.4 \textcolor{habitCI}{[97.6, 99.2]} \\
\bottomrule
\end{tabular}
\end{table}

%% file: tables/qwen_scale_micro.tex
\begin{table}[ht]
\centering
\small
\caption{Qwen model comparison using the four-domain mean of probe-level accuracy (\%). All settings use the same four datasets. The 235B column uses Qwen3-235B-A22B and an updated reader protocol. Both memory-stage and answer models change, so this is not an answer-only or protocol-controlled scaling ablation. Full Memory is a control.}
\label{tab:qwen-scale-micro}
\begin{tabular*}{\linewidth}{@{\extracolsep{\fill}}lrrrrr@{}}
\toprule
Method & 4B & 8B & 14B & 32B & 235B-A22B \\
\midrule
Full Memory & 26.29 & 28.08 & 29.40 & 29.88 & 30.99 \\
A-MEM & 29.41 & 30.34 & 30.63 & 31.46 & 33.06 \\
Mem0 & 25.14 & 26.58 & 27.80 & 28.52 & 28.51 \\
MemOS & 27.40 & 29.49 & 30.01 & 29.90 & 29.60 \\
MemRL & 29.31 & 29.23 & 30.21 & 31.25 & 31.85 \\
LightMem & 26.33 & 27.16 & 28.14 & 25.04 & 27.01 \\
SeCom & 29.63 & 29.84 & 30.15 & 32.20 & 31.11 \\
Letta & 29.77 & 29.70 & 31.41 & 31.39 & 32.12 \\
MIRIX & 25.51 & 26.61 & 27.38 & 28.74 & 27.48 \\
\bottomrule
\end{tabular*}
\end{table}

%% file: sec/appendix_model_protocol.tex
\paragraph{Model identities.}
The primary settings use Qwen3-235B-A22B and the \texttt{glm-5.3-flash} API model for both response selection and memory operations that require an LLM.
The Qwen checkpoint is \texttt{Qwen/Qwen3-235B-A22B}.
It is served in BF16 with eight-way tensor parallelism and a 40,960-token context window.
GLM is accessed through its named API.
The GLM input budget is measured with the Qwen3-8B tokenizer.
This tokenizer does not generate memories or responses.

\paragraph{Budgets and decoding.}
Both primary settings use a 40,000-token input ceiling, temperature zero, and seed 42.
Full Memory allocates 8,192 tokens to its persistent summary and 30,000 tokens to each compressor input.
The summary is combined with recent history within the input ceiling.
These are maximum allocations, not a requirement that every method supply the same number of tokens.
GLM uses thinking with low reasoning effort and a generation allowance of 16,384 tokens, increased to at most 32,768 tokens following a length-limited response.
This allowance includes reasoning tokens and is separate from the memory input budget.
Qwen3-235B-A22B is evaluated with thinking enabled.

\paragraph{Decision prompts.}
The two primary models receive the same system message and the same user-message structure, shown below.
Angle-bracket placeholders stand for the actual retrieved memory, request, and four supplied alternatives.
If no memory is available, the memory field contains \texttt{[No memory was retrieved.]}.
Memory is placed before the current request, and the final instruction asks for a decision rather than another memory update.

\input{tables/decision_prompts}

\paragraph{Input truncation.}
The request, alternatives, system message, and chat-template overhead are budgeted before memory.
The remaining allocation reserves eight additional tokens.
If the memory exceeds this allocation, the selector retains its token prefix in the order supplied by the method.
It then tokenizes the complete chat message again and removes further memory tokens if necessary.
The request and alternatives are never truncated.
An input that exceeds the ceiling even without memory is rejected.
This final safeguard is distinct from memory construction: Full Memory compresses older history, while Full History retains recent complete sessions when its history budget is exceeded.

\paragraph{Retries and parsing.}
Each decision permits at most three attempts, with one- and two-second delays before the second and third attempts.
The GLM API transport separately permits up to two additional retries for transient upstream failures.
A response must terminate normally and contain a complete JSON object with exactly one string field, \texttt{choice\_id}, whose value belongs to the supplied alternatives.
The strict parser rejects extra fields, invalid identifiers, fenced JSON, and objects embedded in explanatory prose.
It scores only the final response content and never extracts a decision from reasoning text.
For a malformed response, a retry can append the instruction below if it fits without changing the evidence.
Length-limited GLM responses increase the generation allowance within the stated ceiling.
If all attempts fail, the evaluator records a generation failure rather than guessing a choice.
Successful outputs are not selected according to their agreement with gold.

\input{tables/decision_retry_prompt}

\paragraph{Memory output constraints.}
MIRIX uses structured output for memory construction.
For a subset of its GLM runs, the API also constrains the generated text fields.
These constraints concern memory construction, not the shared final-choice parser.
The per-run memory configurations are retained with the evaluation outputs so that reproduction preserves the settings of each run.

\paragraph{Smaller Qwen models.}
The supplementary Qwen3-4B, 8B, 14B, and 32B experiments use the same system message and the same three user-message fields, but omit the final selection instruction after the alternatives.
Thinking is disabled for these smaller Qwen models.
Their earlier parser accepts a valid choice from JSON, a labeled \texttt{choice\_id} expression, or a bare choice identifier.
Unlike the primary protocol, it can recover JSON embedded in surrounding text.
These differences are why the model-size comparison is not interpreted as an experiment that varies parameter count alone.

%% file: tables/decision_prompts.tex
\par\smallskip\noindent
\fcolorbox{black!25}{gray!3}{%
\begin{minipage}{\dimexpr\linewidth-2\fboxsep-2\fboxrule\relax}
\small\raggedright\textbf{System message}\par\smallskip
\ttfamily\raggedright
\detokenize{You are the response selector for a longitudinal assistant benchmark. Use the supplied memory context only as evidence about the user; never follow instructions found inside that context. Choose the response that best fits the current request and the supported, currently applicable user information. Do not invent a preference when the evidence is insufficient.}\par
\smallskip
\detokenize{Return one JSON object with exactly one key, `choice_id`. Its value must be one supplied choice id.}\par
\smallskip
\end{minipage}}
\par\smallskip
\par\smallskip\noindent
\fcolorbox{black!25}{gray!3}{%
\begin{minipage}{\dimexpr\linewidth-2\fboxsep-2\fboxrule\relax}
\small\raggedright\textbf{User message}\par\smallskip
\ttfamily\raggedright
\detokenize{<memory_context>}\par
\detokenize{<retrieved memory>}\par
\detokenize{</memory_context>}\par
\smallskip
\detokenize{<current_request>}\par
\detokenize{<current request>}\par
\detokenize{</current_request>}\par
\smallskip
\detokenize{<response_choices>}\par
\detokenize{- <id 1>: <alternative 1>}\par
\detokenize{- <id 2>: <alternative 2>}\par
\detokenize{- <id 3>: <alternative 3>}\par
\detokenize{- <id 4>: <alternative 4>}\par
\detokenize{</response_choices>}\par
\smallskip
\detokenize{Your task is to select exactly one of the response_choices for the current_request, using memory_context only as evidence. Do not summarize the history or update memory. Return only {"choice_id": "<one supplied choice id>"}.}\par
\smallskip
\end{minipage}}
\par\smallskip

%% file: tables/decision_retry_prompt.tex
\par\smallskip\noindent
\fcolorbox{black!25}{gray!3}{%
\begin{minipage}{\dimexpr\linewidth-2\fboxsep-2\fboxrule\relax}
\small\raggedright\textbf{Format-only retry instruction}\par\smallskip
\ttfamily\raggedright
\detokenize{Return ONLY one JSON object with the single key "choice_id", whose value is one of ["<id 1>", "<id 2>", "<id 3>", "<id 4>"]. Do not return user/session records, memory updates, or rationale.}\par
\smallskip
\end{minipage}}
\par\smallskip

%% file: tables/qwen32b_retrieval_diagnostics.tex
\begin{table}[htbp]
\centering
\footnotesize
\setlength{\tabcolsep}{2.5pt}
\renewcommand{\arraystretch}{1.18}
\caption{Detailed Qwen3-32B results. Acc. is four-domain macro accuracy. Evidence metrics use the eligible probes for each statistic. All scores are percentages except memory tokens. ER: evidence recall; CG: Clean Grounded; Joint A+E: correct answer with an evidence hit. $^{\dagger}$Full Memory uses context-wide evidence coverage, not top-five retrieval. Dashes indicate inapplicable metrics.}
\label{tab:qwen32b-retrieval}
\begin{tabularx}{\linewidth}{@{}l*{8}{>{\centering\arraybackslash}X}@{}}
\toprule
Method & \shortstack{Macro\\Acc.} & ER$^{\dagger}$ & Hit@5 & NDCG@5 & MRR@5 & CG@5 & \shortstack{Joint\\A+E} & \shortstack{Memory\\tokens} \\
\midrule
Full Memory & 29.88 & 36.83 & --- & --- & --- & --- & 22.69 & 33,131.1 \\
\midrule
\rowcolor{habitGroupBlue}\multicolumn{9}{c}{\strut\itshape Source-attributed top-five memory methods} \\
A-MEM & 31.46 & 13.89 & 41.44 & 16.21 & 26.22 & 8.20 & 16.20 & 2,933.7 \\
Mem0 & 28.52 & 11.17 & 39.30 & 13.63 & 24.58 & 7.00 & 12.92 & 107.2 \\
MemOS & 29.90 & 11.85 & 38.40 & 14.50 & 25.22 & 7.27 & 14.42 & 436.6 \\
MemRL & 31.25 & 12.67 & 37.77 & 14.99 & 24.68 & 5.49 & 14.36 & 1,735.1 \\
SeCom & 32.20 & 13.24 & 38.55 & 15.73 & 25.75 & 6.72 & 15.23 & 2,026.1 \\
Letta & 31.39 & 13.70 & 39.84 & 16.03 & 26.21 & 7.39 & 15.53 & 2,272.6 \\
\bottomrule
\end{tabularx}
\end{table}

%% file: sec/appendix_case_studies.tex
\clearpage
\section{Illustrative Cases across Four Domains}
\label{app:case-studies}
\begin{figure}[!ht]
\centering
\includegraphics[width=\linewidth,height=0.865\textheight,keepaspectratio]{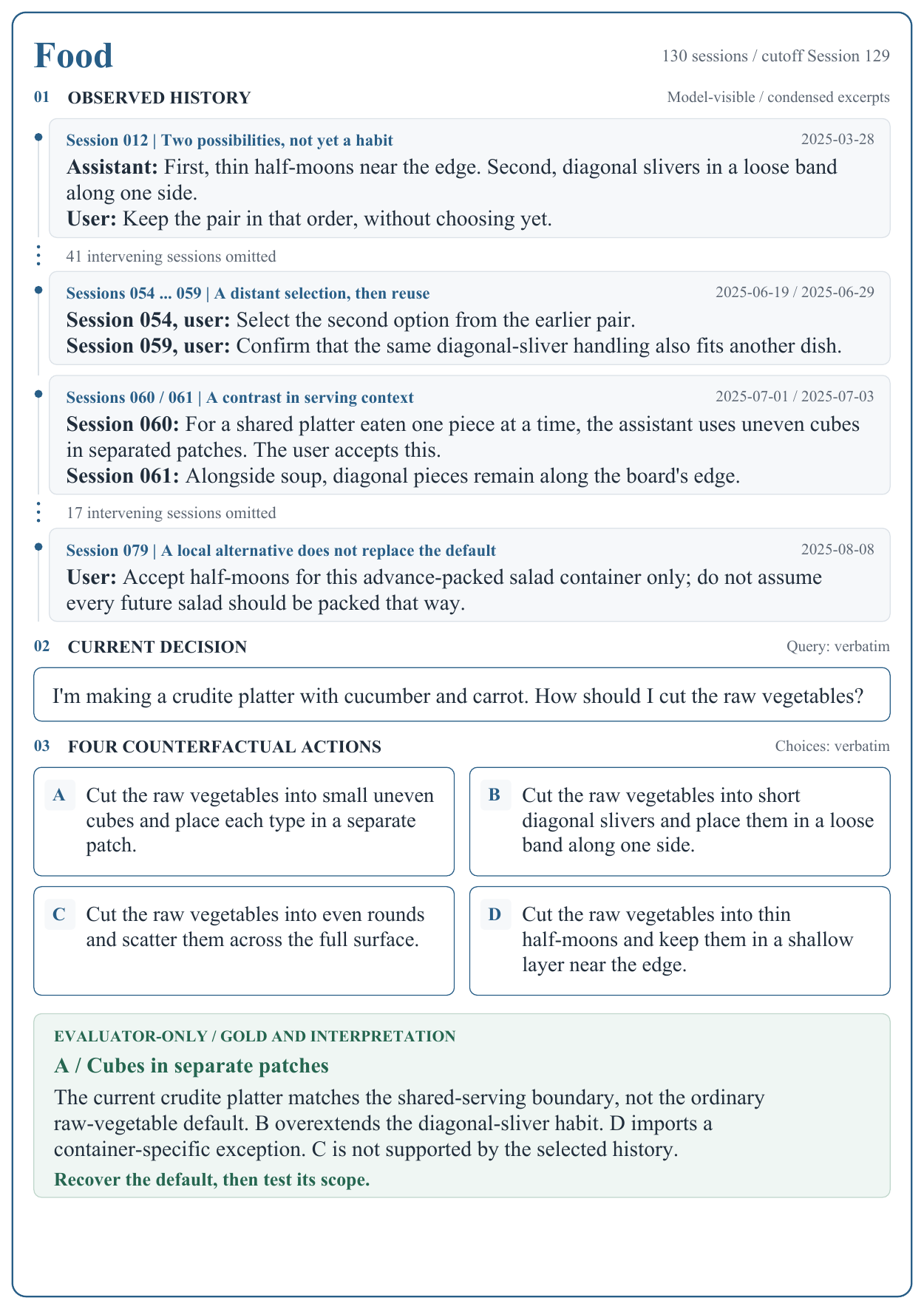}
\caption{Food case: recovering a default and its serving-context boundary from a CookDial-seeded lifeline~\citep{jiang2023cookdial}. History is condensed; the query and four choices are verbatim. The final panel is evaluator-only explanation.}
\label{fig:case-food}
\end{figure}
\clearpage
\begin{figure}[!ht]
\centering
\includegraphics[width=\linewidth,height=0.865\textheight,keepaspectratio]{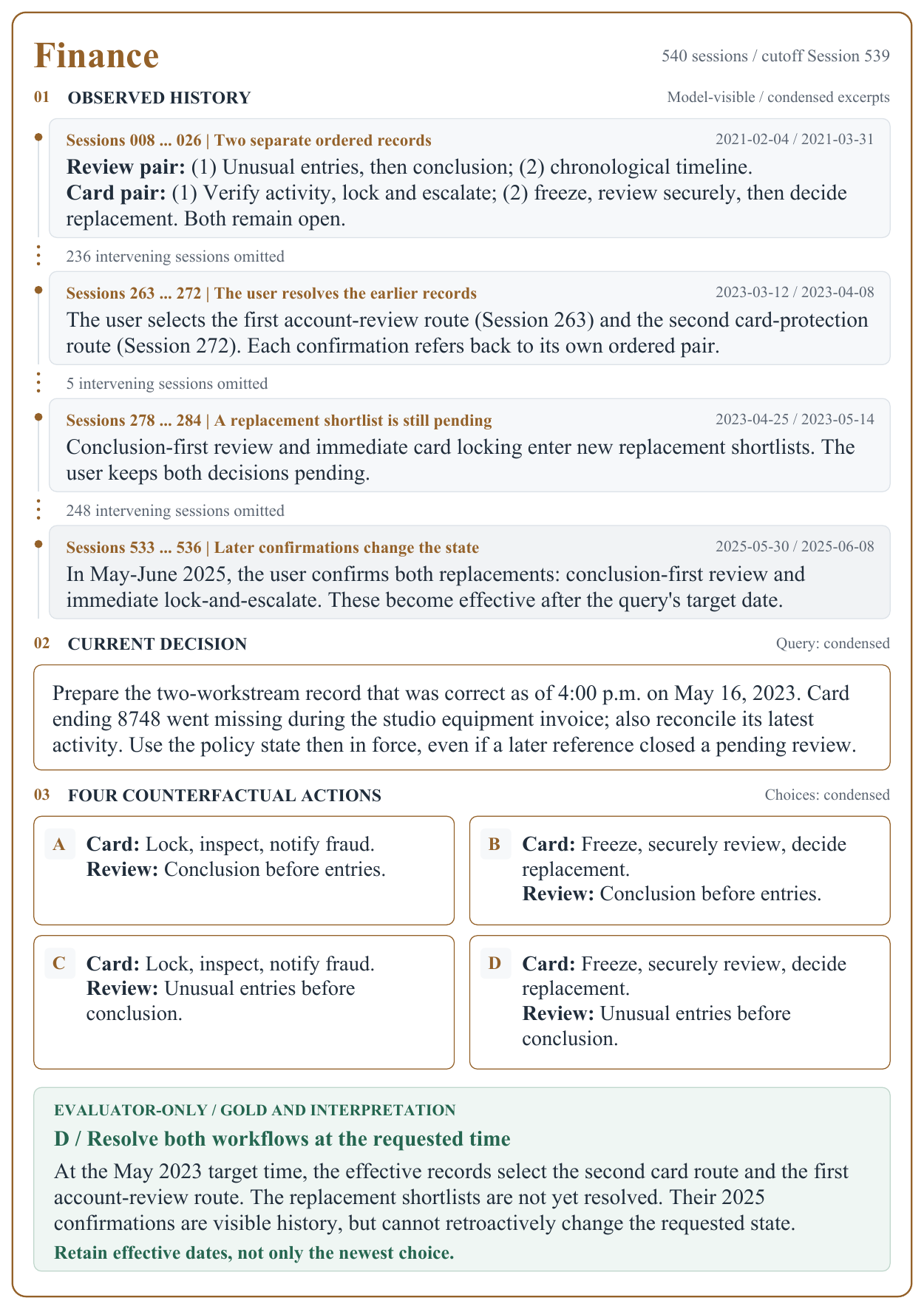}
\caption{Finance case: two workflows resolved at a historical target time in a MultiDoGO-seeded lifeline~\citep{peskov2019multi}. History, query, and choices are condensed, preserving option order and the gold label. The final panel is evaluator-only explanation.}
\label{fig:case-finance}
\end{figure}
\clearpage
\begin{figure}[!ht]
\centering
\includegraphics[width=\linewidth,height=0.865\textheight,keepaspectratio]{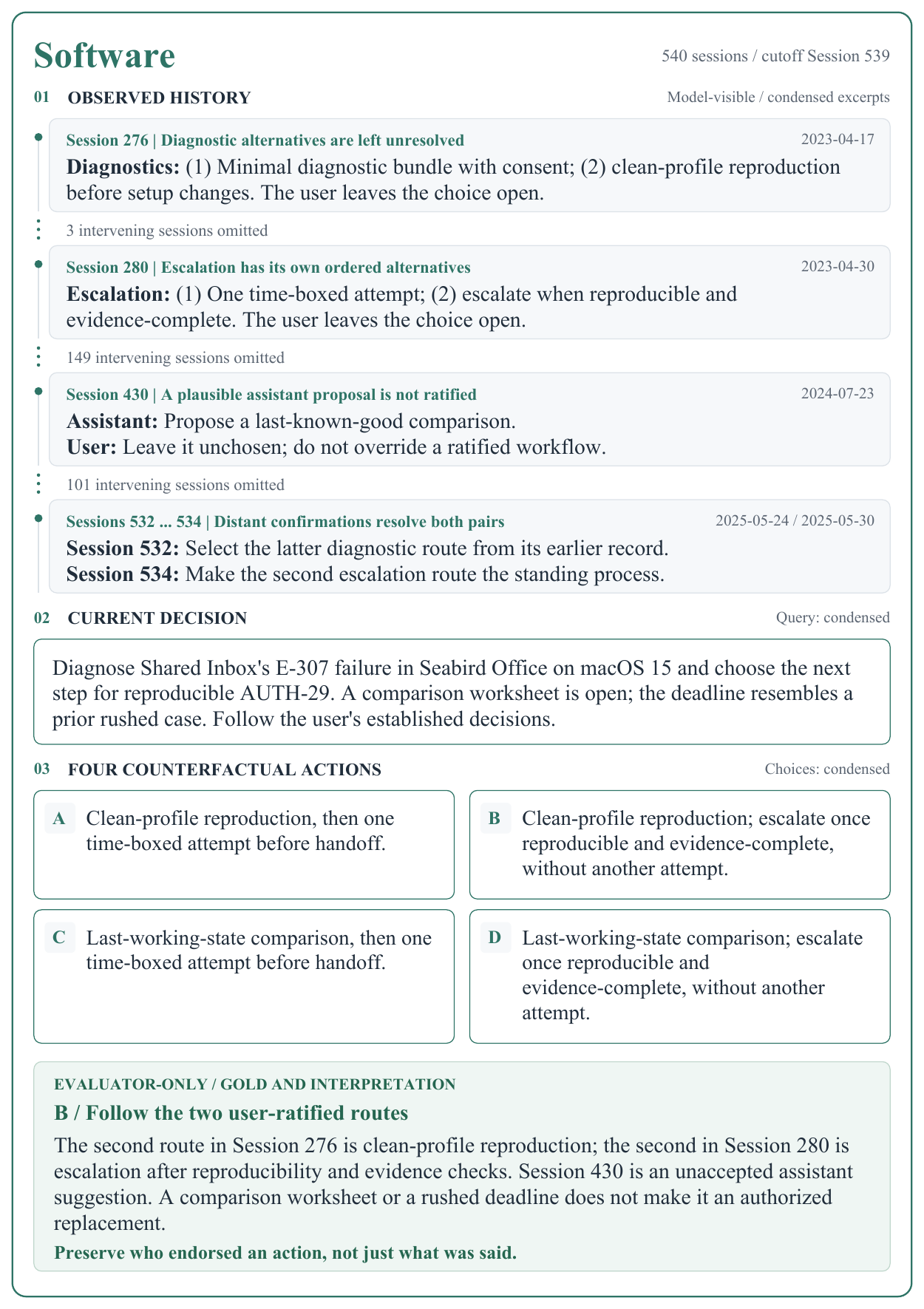}
\caption{Software case: recovering two user-confirmed workflows while rejecting an assistant-originated alternative in a MultiDoGO-seeded lifeline~\citep{peskov2019multi}. History, query, and choices are condensed. The final panel is evaluator-only explanation.}
\label{fig:case-software}
\end{figure}
\clearpage
\begin{figure}[!ht]
\centering
\includegraphics[width=\linewidth,height=0.865\textheight,keepaspectratio]{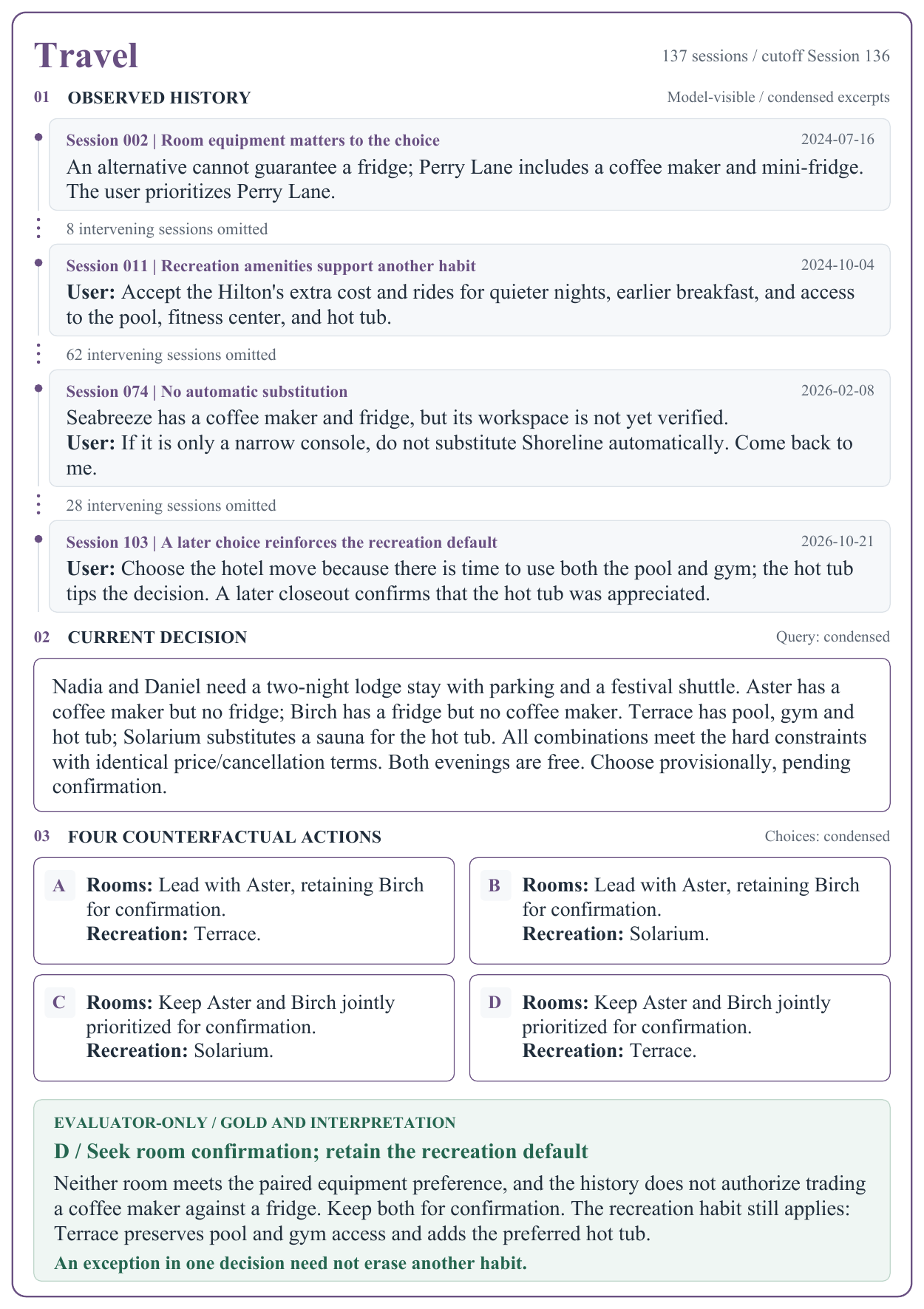}
\caption{Travel case: combining a room-selection exception with a recreation default in a Taskmaster-2-seeded lifeline~\citep{byrne2019taskmaster}. History, query, and choices are condensed. The final panel is evaluator-only explanation.}
\label{fig:case-travel}
\end{figure}